\documentclass[a4paper]{article}

\usepackage[pages=all, color=black, position={current page.south}, placement=bottom, scale=1, opacity=1, vshift=5mm]{background}
\SetBgContents{
	\tt This work is shared under a \href{https://creativecommons.org/licenses/by-sa/4.0/}{CC BY-SA 4.0 license} unless otherwise noted
}      

\usepackage[margin=1in]{geometry} 

\usepackage{amsmath}
\usepackage{amsthm}
\usepackage{amssymb}

\usepackage[utf8]{inputenc}
\usepackage{hyperref}
\hypersetup{
	unicode,
	pdfauthor={Author One, Author Two, Author Three},
	pdftitle={A simple article template},
	pdfsubject={A simple article template},
	pdfkeywords={article, template, simple},
	pdfproducer={LaTeX},
	pdfcreator={pdflatex}
}

\usepackage[sort&compress,numbers,square]{natbib}
\theoremstyle{plain}

\theoremstyle{definition}

\usepackage{graphicx, color}
\graphicspath{{fig/}}

\usepackage{algorithm, algpseudocode} 
\usepackage{mathrsfs} 

\usepackage{lipsum}

\usepackage{lettrine}
\usepackage{multicol}
\usepackage{multirow}
\usepackage{pifont}
\usepackage{caption}
\usepackage{subcaption}
\usepackage{color}
\usepackage{booktabs} 
\newcommand{\cmark}{\ding{51}}
\newcommand{\xmark}{\ding{55}}

\title{\textbf{Neuromorphic Perception and Navigation for Mobile Robots: A Review}}
\author{A. Novo$^a$\thanks{Corresponding author \\Email address: martineznovo@ugr.es}, F. Lobon$^b$, H.G. De Marina$^a$, S. Romero$^a$, F. Barranco$^a$}

\begin{document}
    \date{\vspace{-5ex}}
	\maketitle
    \begin{center}
	{\small
	$^a$Department of Computer Engineering, Automation and Robotics, Research Centre for Information and Communication Technologies (CITIC-UGR), University of Granada, 18014, Granada, Spain\\%
	$^b$Instituto de Astrofísica de Andalucía (IAA-CSIC), 18008, Granada, Spain \\%
    }
    \end{center}
    \hrule
	\begin{abstract}
		With the fast and unstoppable evolution of robotics and artificial intelligence, effective autonomous navigation in real-world scenarios has become one of the most pressing challenges in the literature. However, demanding requirements, such as real-time operation, energy and computational efficiency, robustness, and reliability, make most current solutions unsuitable for real-world challenges. Thus, researchers are forced to seek innovative approaches, such as bio-inspired solutions. Indeed, animals have the intrinsic ability to efficiently perceive, understand, and navigate their unstructured surroundings. To do so, they exploit self-motion cues, proprioception, and visual flow in a cognitive process to map their environment and locate themselves within it. Computational neuroscientists aim to answer ``how'' and ``why'' such cognitive processes occur in the brain, to design novel neuromorphic sensors and methods that imitate biological processing. This survey aims to comprehensively review the application of brain-inspired strategies to autonomous navigation, considering: neuromorphic perception and asynchronous event processing, energy-efficient and adaptive learning, or the imitation of the working principles of brain areas that play a crucial role in navigation such as the hippocampus or the entorhinal cortex.
		
		\vspace{1ex}\noindent\textbf{Keywords:} navigation, hippocampus, neuromorphic sensors, brain-inspired
	\end{abstract}

\section{Introduction}
\label{sec:intro}
\lettrine[findent=2pt]{\textbf{R}}{\textbf{obotics}}\textbf{, more than an engineering, is an Art.} From the earliest robotics breakthroughs, nature has been a fundamental inspiration to pursue the paradigm of the reliable autonomous system. Bio-mimetic mechanical designs \cite{Thomas_2014}, neuromorphic sensors \cite{Lichtsteiner_2008}, swarm behavior policies \cite{dorigo_2021}, or even algorithms that imitate the behavior of brain regions \cite{Fangwen_2019} are the focus of intensive research in robotics. Autonomous robots perform their tasks within a certain level of self-sufficiency, avoiding, to some extent, human intervention \cite{machines5010006}. Most early autonomous robots required an on-purpose configuration of their environment, with e.g. markers or beacons \cite{Ritz_2012}. Such a setup constrains the robots to a specific environment, with a costly preparation. Today, autonomous robots are asked to interact with dynamic and unstructured environments \cite{Tordesillas_2022}, typically without following a fixed script. Indeed, current autonomous robots perceive their environment to sense, plan, communicate, and finally actuate on their environment, usually requiring changes on it \cite{correll2022introduction}. 

The novel field of neuromorphic engineering investigates and exploits the potential of sensors, computing platforms, and algorithms attempting to mimic biological capabilities. This area emerges in an attempt to achieve what nature proves: how animals effectively perceive, plan, and act, in an energy-efficient and robust fashion and adapt to changing circumstances in real-time. All these properties are of great value for robotics. Attention has also been drawn toward how the brain internally works to accomplish navigation tasks, the brain regions involved in such processing, and their working principles. Although multiple areas are needed, the process mainly occurs in the hippocampus and the entorhinal cortex areas. These areas make possible positioning in a map of the environment with minimal effort \citep{samerican2016}. Also, they enable path integration for autonomous navigation. Taking motion cues such as the direction and speed of the individual over time, path integration is able to estimate its position within the map of the scenario. Compared to traditional methods where this process is computationally expensive, our brain does it with minimal resources in such a way that we are not even conscious of that. This is thanks to the cognitive map that our brain internally builds when certain groups of cells produce a firing pattern that encodes our location within that map \citep{Bermudez2020}. From this cognitive map, experiments with rats proved that they are able to take shortcuts when navigating, showing that this mental map representation does not only take geometrical information of the outer world, but they also appeared to record information about the events that the animals experienced at specific locations. This capability may be useful in building a new paradigm for the well-known traditional planning methods, where the input mapping representation is crucial for deriving real-time optimal paths based on events for autonomous robots.

\textbf{Motivation:}
Neuromorphic devices that mimic biological neural systems at different levels like sensing, communication protocols, architectures, or strategies have been the focus of intense research for decades \cite{vanarse2019neuromorphic}. On the one hand, engineers have taken inspiration from how the brain works and solves very complex problems in the real world to develop new algorithms and devices. On the other hand, neuroscientists have tried to understand the functionality of the different neural circuits in the brain. Despite the great advances in the last decades, the promise of a massively parallel model based on simple computational units for low-power and real-time computing that mimics biological neural systems with state-of-the-art accuracy remains unfulfilled \citep{Dampfhoffer_2023}\cite{Schuman2022}. The review scope is limited to vision-based asynchronous event processing since they are the most popular solutions thanks to the wide availability of neuromorphic vision sensors. 

Precisely, the main purpose of this survey is to review how the scientific community faces the problem of autonomous navigation and to characterize the existing solutions according to brain-inspired principles. Moreover, we are committed to shed light on their potential applications, discussing the advantages and limitations of these solutions when used in the real world, compared to conventional approaches. Each robotic navigation subsystem (i.e. localization, mapping, or planning) is individually discussed, with a focus on works that are purely bio-inspired. The objective of this article is to respond to the question \textit{``Could neuromorphic engineering solutions succeed the current and classical navigation approaches in robotics?''}

\textbf{Outline:}
Next, we describe the representation and processing of sensory information from a biological perspective in Section \ref{sec:methods}. In Section \ref{sec:perception}, sensors based on neuromorphic vision, commonly known as \textit{event cameras}, are described. The core of this review is addressed in Section \ref{sec:visual_navigation}, which covers bio-inspired navigation methods, presenting the whole autonomous navigation pipeline in separate subsystems that are distributed in subsections. A final conceptualized flow diagram of this autonomous pipeline can be found in Fig. \ref{fig:diagram}. Then, Section \ref{subsec:brain_spatialcells} discusses some specific biological approaches based on brain areas known to be involved in navigation. Finally, we present the discussion and conclusions in Section \ref{sec:discussion} and Section \ref{sec:conclusions}.

\begin{figure}[!t]
    \centering
    \includegraphics[width=0.9\textwidth]{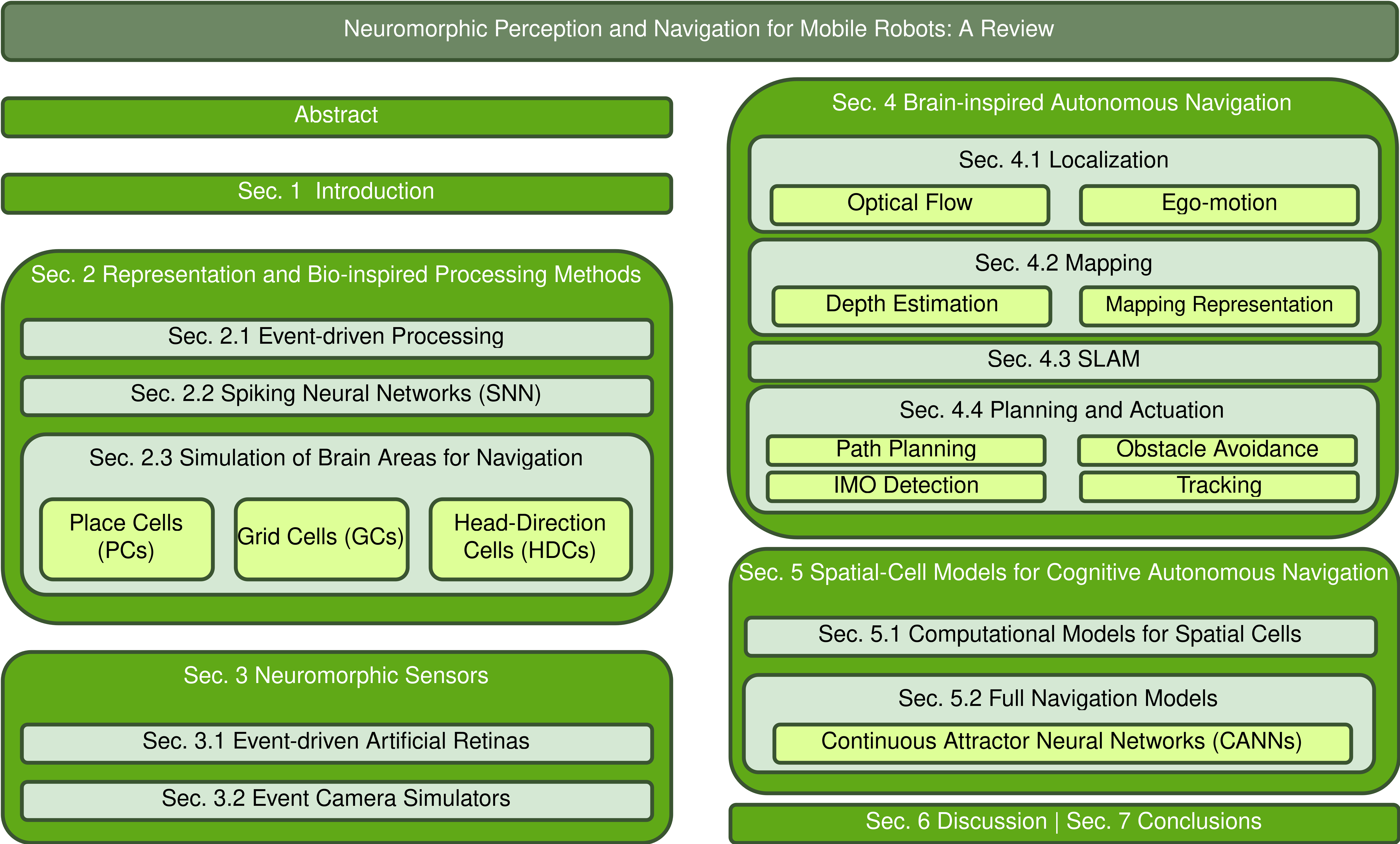}
    \vspace{-2.5mm}
    \caption{Structural and hierarchical taxonomy of the survey.}
    \label{fig:survey_recap_diagram}
    \vspace{-3.5mm}
\end{figure}

\section{Representation and Bio-Inspired Processing Methods} 
\label{sec:methods}
Traditional approaches based on mathematical models designed from specific domain knowledge are very efficient due to their simplicity; typically, they avoid the need for large amounts of data as required in Machine Learning approaches. However, these traditional models use statistical or geometrical methods based on measurements of a deterministic domain representation and thus, do fail with high-dimensional dynamic problems \cite{shlezinger2023model}. In this survey, these domain-knowledge-based methods are simply referred to as \textit{model-based approaches} as opposed to \textit{learning-based approaches}. However, this survey is focused on brain-inspired methods and thus, we pay special attention to strategies that use neuromorphic data: event-driven processing, adaptive and robust learning with Spiking Neural Networks, and models that imitate the way in which some of the main brain areas involved in navigation work. 

\subsection{Event-driven processing}
State-of-the-art visual processing is based on intensive computing fed with sequences of synchronous images. Differently, biological solutions use continuous streams of asynchronous events that are processed by hierarchical structures to extract features from the scene. This asynchronous independent processing is called event-driven and enables real-time and low-power computation \cite{camunas2014event}.     

Events are triggered when a certain threshold is crossed and are emitted as a tuple with 2 fields, using the Address-Event Representation (AER): a) the event \textit{address} corresponds to the coordinates within the sensor matrix that triggered it and b) the \textit{time} of event creation. AER is robust to aliasing, potentially parallelizable, and composable with the ability to easily build hierarchical structures \cite{tayarani2021event}. Unlike frame-based representations, the acquisition is triggered by spatio-temporal changes, and thus, the captured information better represents scene dynamics.

Most approaches to processing neuromorphic data often use well-known solutions that adapt conventional methods. From an engineering point of view, we believe these solutions are valuable if they represent an advancement in the field without ignoring the benefits of event data. However, approaches that merely transform events into artificial frames to be processed by conventional methods, ignoring the scene dynamics, do not take full advantage of the high-temporal resolutions of event cameras. Moreover, building artificial frames from events yields processing redundant information, increasing the required computing resources and data bandwidth.

\subsection{Spiking Neural Networks}
SNNs are the learning counterpart in neuromorphic computing to conventional neural networks, also known as Artificial Neural Networks (ANNs). ANNs, and specifically the successful DNNs (Deep Neural Networks), have become very popular to solve problems in all different fields. However, the metric of success has exclusively been focused on accuracy, neglecting other qualities such as computational complexity, energy consumption, or storage needs \cite{davidson2021comparison}. SNNs emerged to model the behavior of biological neuron networks, where the neuron usually integrates its input spike voltages (action potentials), firing a new spike when the accumulated voltage crosses a threshold (membrane potential). Information is rate- or time-coded via spikes, making transfer fault-tolerant, meaning that missing an individual spike does not significantly impact the average firing rate. Also, given the sparsity of the inputs, SNNs only process information when there are spikes, resulting in reduced computational load and energy consumption compared to traditional continuous computation in ANNs.

\begin{figure}[!t]
	\centering
	\includegraphics[width=0.9\textwidth]{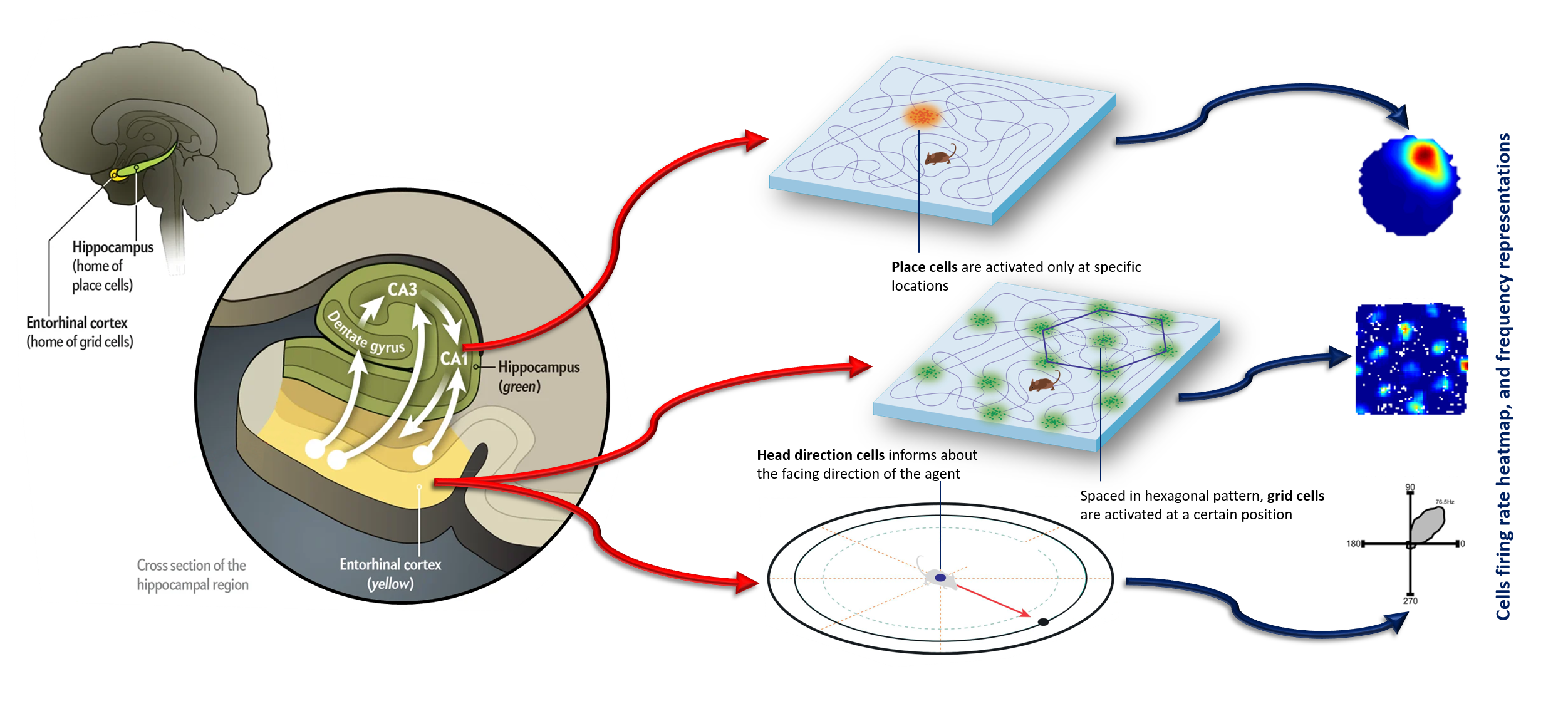}
        \vspace{-4.5mm}
	\caption{The underlying brain navigation cells process (adapted from \cite{Grieves_2017}, \citep{samerican2016}, and \citep{frontiers2021}).}
	\label{fig:LIF_model_and_cells}
        \vspace{-3.5mm}
\end{figure}

SNNs have proven their potential to achieve precision results on pair with ANNs while reducing computational cost and energy consumption \cite{davidson2021comparison}. However, their main drawback is the lack of a learning rule similar to backpropagation in the powerful supervised learning in ANNs, which enables efficient training. To apply backpropagation for gradient-descent methods, continuous differentiability is required (spikes are not). Some approaches to overcome this limitation are: 1) to transfer weight values from ANNs to SNNs with similar architectures \cite{diehl2015fast}; 2) to adapt backpropagation to SNNs, e.g. using Surrogate Gradient Learning \cite{neftci2019surrogate} achieving great accuracy gains; 3) or to train using bio-inspired rules such as Spike-Timing-Dependent Plasticity (STDP), a rule that excites the synapse connecting two neurons if their triggers are correlated and inhibits them otherwise, enabling real-time adaptive learning \cite{diehl2015unsupervised}.

Finally, encouraging hardware development in the last years has made SNNs the focus of interest for the industry, with platforms such as Intel Loihi \cite{davies2018loihi}, IBM TrueNorth \cite{akopyan2015truenorth}, or Brainchip Akida \cite{vanarse2019hardware}.

\subsection{Simulation of brain areas for navigation}
\label{subsec:simulation}
Several areas in the brain are known to be involved in cognitive processes regarding navigation, being the most relevant ones the hippocampus and the entorhinal cortex. These areas contribute to building cognitive map representations of the spatial environment or the location of the individual in that map \cite{Grieves_2017, whittington2022build}. Neuroscientists point out that navigation is possible thanks to electrophysiological correlates of various types of cells in the brain concretely, three main types that contribute to the brain spatial representation and navigation support \cite{moser2015place}:

\begin{itemize}
    \item Place Cells (PC): They contribute to building a cognitive representation of the space. These cells maximize their firing rate when the animal is in one specific region, namely \textit{Place Field}, regardless of its orientation and what it sees \cite{Jeffery_2015}. PCs are thought to be involved in recognizing a previously visited scenario and in route planning \cite{Miao_2015}. 
    
    \item Grid Cells (GC): These cells are thought to cooperate for self-motion calculation, which is indeed part of the path integration process. Also, they are thought to be part of the pose estimation process while the agent navigates \cite{moser2008metric}.
    
    \item Head Direction Cells (HDC): These cells fire for a specific orientation, maximizing their firing rate at a \textit{preferred direction} independently of the agent position. These cells are believed to fully encode the pose of the agent in the environment \cite{Taube_1990}. 
\end{itemize}

Regarding the environment representation, space can be represented as 1) an \textit{egocentric} map where space is defined concerning the agent itself, 2) an \textit{allocentric} representation defined by the relationship between external space features captured during navigation, or 3) an \textit{inertial} map where location and orientation are defined relative to the direction and distance traveled from an initial reference point. The discovery of HDCs led to the theory that these cells are doing integration of head angular velocity, and thus that these brain areas perform \textit{path integration} to build their internal coordinate system with the additional involvement of GCs. In other words, \textit{path integration} is performed using egomotion cues such as optical flow and vestibular signals, to reconstruct the relative pose and distance traveled from the starting point \cite{moser2017spatial}. Moreover, \textit{place recognition} is another critical cognitive process with experiments that suggest that visual information is merged at the level of the entorhinal cortex (EC) \citep{Gaussier2019} as a cognitive map. PCs are believed to do so by merging the ``what'' and ``where'' information related to visual stimuli, even when the environment changes dramatically. Eventually, this last is a key feature for Simultaneous Localization and Mapping (SLAM) algorithms in autonomous robotics. Refer to Section \ref{subsec:brain_spatialcells} for a detailed review of the computational models that replicate the working principles of these brain areas. 

\section{Neuromorphic Sensors}
\label{sec:perception}
Neuromorphic sensors are a type of electronic circuit design that mimics biology, in an attempt to take advantage of the efficient cognition processes evolved to improve their performance for millions of years. Applying this so-called neuromorphic engineering to machines opens up new potential to improve their efficiency, latency, or power consumption. 

During the last decades, many neuromorphic sensors have been developed: silicon cochleas that convert vibrations in the air into event-driven signals used, e.g., for sound localization \cite{liu2014event}; olfactory sensors that identify volatile chemical components in the air \cite{chicca2014neuromorphic}; or spiking tactile sensors that consist of mechanoreceptors that perceive physical distortion and thus shapes or texture, with applications in robotic grasping or prosthetics \cite{birkoben2020spiking}. However, given the dominance of visual perception \cite{kandel2000principles} most works in the literature are devoted to event-driven artificial retinas.

\subsection{Event-driven artificial retinas}
\label{subsec:retinas}
Event cameras are imaging sensors that do not capture synchronous frames as conventional cameras do. Every single pixel of the event sensor responds independently to the luminance of the scene, triggering events asynchronously when significant changes occur. Moreover, since each pixel is processed independently, the temporal resolution is very high, with latencies in the order of a few microseconds. In some manner, event cameras naturally perform a smart compression of the scene triggering events only for moving edges in the scene, at a very high speed. All these features hint at the potential of event cameras as very promising sensors for navigation tasks \citep{Gallego_2022}.  

\begin{figure}[!t]
	\centering
	\includegraphics[width=0.9\textwidth]{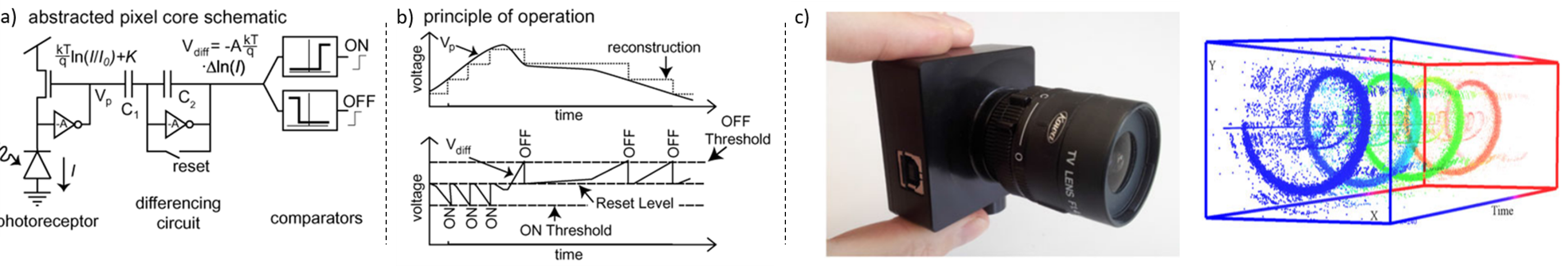}
        \vspace{-2.5mm}
	\caption{a) Event camera pixel schematic, and b) principle of operation (from \cite{Lichtsteiner_2008}). c) Time-space accumulation of events triggered by an event camera for a spinning object.}
	\label{fig_0}
        \vspace{-2.5mm}
\end{figure}

In detail, for each pixel, a reference log intensity level $L \approx log(I)$ is stored and compared to the current value at every new timestamp. If the difference goes over a predefined threshold $\Delta L = C$, an event is triggered for this pixel location as a tuple $(\textbf{x},t,p)$, where $\textbf{x}$ corresponds to the 2D pixel location, $t$ is the timestamp at which the event was triggered, and $p$ the polarity or sign of the log intensity change. Eq. \ref{eq:event_cameras} shows this working principle, where $\nabla L$ is the log intensity gradient.
\vspace{-1.5mm}
\begin{equation}
- \langle \nabla L, \textbf{x}\Delta t \rangle = C
\label{eq:event_cameras}
\vspace{-1.5mm}
\end{equation}
Briefly, the output is an asynchronous stream of events triggered by the relative changes in the scene luminance at a very high speed. The main features of event cameras are:
\begin{itemize} 
\item \textbf{High dynamic range (HDR):} Event cameras have an HDR of up to 120~dB (conventional cameras over 60~dB). This is an advantage under challenging lighting conditions such as low illumination or sudden contrast, as biological retinas do. In navigation, this is relevant to reduce artifacts when e.g. driving out of a tunnel or entering a room.  
\item \textbf{Low latency:} Contrarily to conventional cameras which need to wait for a global exposure time to capture a frame, each pixel value of an event camera is independently processed and transmitted with sub-millisecond latencies. Thanks to this property, event cameras are not severely impacted by motion blur.
\item \textbf{Low-power consumption:} Since event cameras do only capture luminance changes, areas with redundant static information do not trigger any new events. This usually leads to neuromorphic devices consuming less energy (in the order of a few microwatts vs. tens of milliwatts for conventional sensors). This is another strong point for using this technology in autonomous robots with limited battery power.
\end{itemize}

Finally, event-based sensors are well-suited for applications where dynamics play a significant role, as happened in navigation. As shown in \cite{deniz_2023}, conventional approaches based on full frames mostly extract spatial features and need to reconstruct the dynamics from the sequences of frames whereas event-driven approaches are the best fit for extracting dynamic spatiotemporal features, which are the key features for navigation processing. Despite all these advantages, the sparsity of event inputs makes effective event-based navigation still a challenging open problem. 

In addition, some of the most prominent hardware companies that specialize in event sensor manufacturing are \textbf{iniVation} with their popular series named DAVIS sensors \citep{davis}, \textbf{Samsung} with their neuromorphic DVS sensors \citep{Suh2020A1D}, and \textbf{Prophesee} and their ATIS sensors \citep{Finateu2020510A1}.

\subsection{Event camera simulators}
\label{subsec:simulators}
With the success of event cameras and the adoption of deep learning solutions, new large datasets are needed for the advancement of the field. Moreover, event cameras are expensive, and commercial systems are still prototypes, suffering from limitations in their resolution or interfaces. These are the main reasons for building event camera simulators that face challenges in efficiently obtaining realistic data.     

ESIM \cite{rebecq2018esim} was one of the first complete simulators released as an open platform. This simulator used adaptive rendering to simulate the asynchronous nature of event cameras, providing large amounts of event data. The validation is done by training a network for optical flow estimation with simulated data and evaluating its generalization capabilities when inferring real events. Another very popular simulator is V2E \cite{hu2021v2e}, which uses a similar methodology for validation but focuses on the realism of their simulated events. V2E uses a deep learning solution for video interpolation to increase the temporal resolution of the source videos. Both simulators were also compared and extended into the ICNS model \citep{Joubert2021}, which notably improves the noise simulation, and it is also validated with a real sensor. In \citep{snyder2023object}, the ViBES simulator was used to mimic the Object-Motion-Sensitivity (OMS) function of the amacrine and retinal ganglion cell layers \citep{Baccus2008ARC, Yin2023}. Its goal is to reduce the dimensionality of visual information in each cell to a more encoded feature-rich representation. Finally, \citep{Zhao2022} presented SpikingSIM, a simulator that generates spiking data from images recorded by conventional frame cameras, including realistic spike noise.

\begin{figure}[!t]
    \centering
    \includegraphics[width=0.9\textwidth]{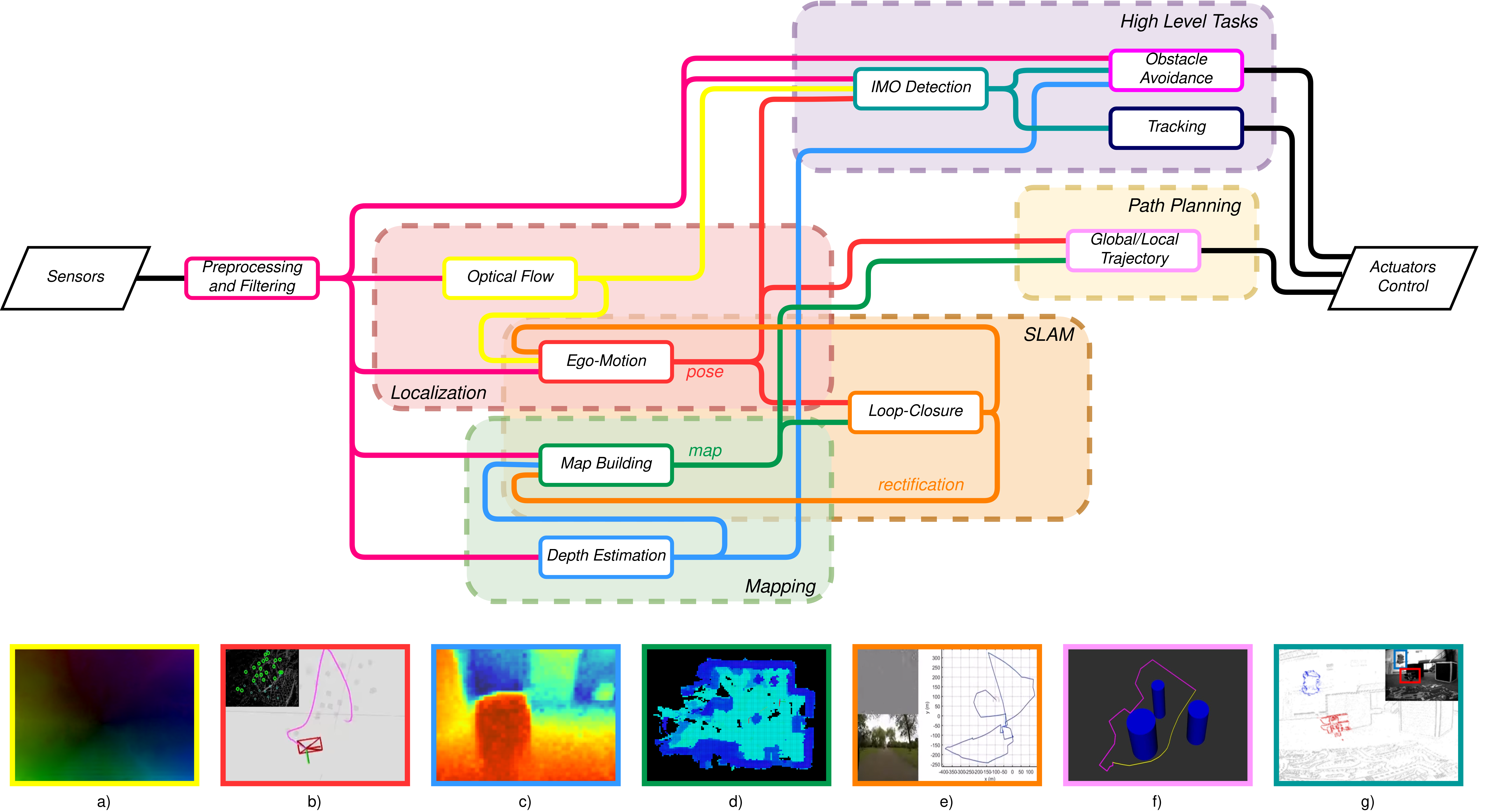}
    \vspace{-2.5mm}
    \caption{Autonomous navigation flow diagram. Top: For each subsystem, the left arrows are inputs and the right ones are outputs. Bottom: examples of subsystem outputs: a) Optical flow \cite{Gehrig_2021}, b) Ego-motion \cite{Vidal_2017}, c) Depth \cite{ranccon2021stereospike}, d) Mapping \cite{hornung13auro}, e) SLAM \cite{Milford2015TowardsVS}, f) Path planning \cite{Novo_2022}, g) IMO detection \cite{Parameshwara_2020}.}
    \label{fig:diagram}
    \vspace{-3.5mm}
\end{figure}


\section{Brain-inspired Autonomous Navigation}
\label{sec:visual_navigation}
\textit{Guidance, Navigation, and Control} (GNC) is an engineering branch focused on the motion of, mostly autonomous, vehicles in their environments. In this survey, we limit our interest to the navigation part. Navigation refers to the measurement and estimation of the robot state vector that, depending on the system, is usually composed of the pose and velocity (and sometimes complemented with acceleration, attitude, rotation rates, etc). Navigation usually includes the sensor selection and modeling, the estimation of the deviations concerning the reference trajectory, and the planning and execution of actions to reduce these deviations between the actual and reference trajectories. Autonomous navigation is usually implemented by combining visual sensors, inertial units, laser, LiDAR, or Global Positioning System (GPS). However, we will focus on visual navigation techniques. 

In the next subsections, we cover neuromorphic approaches to build maps and localize agents within them. We also review the most popular path-planning methods, as well as other related tasks such as obstacle avoidance or the detection of independently moving objects (IMOs). Fig. \ref{fig:diagram} shows a flow diagram for the pipeline of autonomous navigation, highlighting the main components and their relationships, and also illustrating the processing with some results from reviewed methods.

\subsection{Localization}
\label{subsec:localization}
Localization enables the continuous tracking of the robot's position and orientation while it moves through the environment, thanks to sensory inputs. However, these sensory inputs are noisy, which leads to incorrect position and orientation estimates and ambiguities. The estimation can be refined by combining additional cues relative to the scene landmarks, the motion of the objects in the scene (optical flow), and the robot's self-motion (ego-motion), which are provided by vision sensors. 

\subsubsection{Optical Flow}
\label{subsubsec:OF}
Optical flow determines each pixel velocity in the 2D image plane. Thus, it can be directly linked to the robot's speed and consequently to its state vector. Traditional optical flow estimation algorithms cannot be applied, at least directly, to event data, as they usually rely on the minimization of the frame-to-frame pixel brightness, also known as brightness or Optical Flow Constraint (OFC). Event cameras do not produce brightness information, but only events when changes in the brightness for the same location reach a certain threshold. However, event information is substantially more valuable, since it encodes the spatiotemporal dynamics of the scene, focusing on edges that tend to be more robust for flow estimation. Furthermore, image-based methods struggle in the presence of artifacts such as saturation or motion blur because of the limited dynamic range of conventional cameras, or their limited temporal resolution. High dynamic range and high temporal resolution are exactly two of the advantages that event-driven methods can exploit if using neuromorphic sensors as mentioned in Section \ref{subsec:retinas}. 

Event model-based approaches can be categorized into frame-based Lucas-Kanade \cite{Benosman_2012, Rueckauer_2016_dataset}, filter-bank-based \cite{barranco2015bio, brebion2021real}, spatio-temporal plane-fitting \cite{Benosman_2014, Rueckauer_2016_dataset, Akolkar_2018}, variational \cite{Bardow_2016}, and contrast maximization \cite{shiba_2022} methods. Apart from them, the most recent approaches are learning-based, using either ANNs or SNNs. The most common architectures are U-Net \cite{Ronneberger_2015}, FireNet \cite{Scheerlinck_2020}, and long-term memory-based networks \cite{Teed_2020}. These learning-based methods are categorized into supervised, self-supervised, or unsupervised approaches.

Next, we describe in detail the most relevant works in the literature for optical flow, distinguishing between model-based methods, deep learning approaches using conventional ANNs, and SNN-based solutions, and summarize the most relevant details in Table \ref{tab:OF_methods}. 

\textbf{Traditional Methods}. The first approximation for optical flow estimation is to build artificial frames to be processed by conventional methods. However, this strategy sacrifices the low latency and high-temporal resolution of asynchronous events with the burden of computing a large number of artificial frames as conventional methods do. In practice, most traditional methods start with the OFC \cite{LKTracker}. Assuming no illumination changes (at time $t$ and $t+1$), the pixel brightness $I$ remains constant. Therefore, the OFC can be written as $I(\textbf{x},t)-I(\textbf{x}+\textbf{u},t+1)=0$, where the motion of pixel $\textbf{x}$ is $\textbf{u}=(u,v)$. Applying the first-order Taylor series expansion, the OFC is expressed as $\nabla I \cdot \textbf{u} = -I_{t}$. Extending it, the equation can be rewritten in its most familiar form $I_{x}u + I_{y}v + I_{t} = 0$,  with $I_x,I_y,I_t$ the spatial and temporal partial derivatives. Since the equation has two unknowns $u, v$ and only one equation, it is usually extended with new constraints to solve it. For example, Horn and Schunck method \cite{horn_80} assumes that the optical flow is spatially smooth, solving for $\textbf{u}$ by minimizing the expression in Eq. \ref{eq:horn}.
\vspace{-1.5mm}
\begin{equation}
    \underset{{\textbf{u}}}{\mathrm{arg\,min}} \int_{\Omega}^{} \lambda^{2}( \left\| \nabla u \right\|^{2} + \left\| \nabla v \right\|^{2}) \,+  
    (\nabla I \cdot \textbf{u} + I_{t})^{2} \text{d}\textbf{x}
    \label{eq:horn}
\vspace{-1.5mm}
\end{equation}
In this expression, the error includes the regularization term which serves to smooth the flow, and the data term in this case is the original OFC, with $\lambda^2$ as the smoothness weighing factor.

\textit{Lucas-Kanade}. In \cite{Benosman_2012} authors proposed the first approach that naively adapts the Lucas-Kanade method to events, approximating spatial and temporal partial derivatives as the derivatives of the accumulation of events at a pixel for a period of time. This approximation generally fails when edges are not sharp or scenes are complex including textures. 

\textit{Local plane-fitting}. Later in \cite{Benosman_2014}, authors presented a new model based on matching local plane fit over time. Fitted spatio-temporal planes overcome the sparsity of neighboring asynchronous events, allowing for more robust estimations. Instead of using temporal windows, the authors use the time from the events for the estimation, becoming also more biologically plausible. However, due to the eminently local nature of this approach, it does not correctly handle the aperture problem. To overcome this issue, \cite{Akolkar_2018} used a multi-scale spatial pooling approach. To do so, they initialized the optical flow using the local plane fitting approach. The authors then applied a multi-scale approach to estimate the refined optical flow, using a contrast-maximization approach being able to handle more complex scenes with general camera motion that involves 3D rotation and translation. 

\textit{Variational}. 
In \cite{Bardow_2016}, authors introduced another method to estimate dense optical flow, reconstructing intensity from events, using a variational approximation (based on Horn-Schunck \cite{horn_80}). The authors proposed to solve for the optical flow $\textbf{u}$ by reconstructing also the log intensity from the events $L$, by minimizing the Eq. \ref{eq:bardow}.
\vspace{-1.5mm}
\begin{equation}
\begin{split}
    \underset{\textbf{u}, \, L}{arg\,min} & \int_{\Omega}^{} \int_{T}^{} \lambda_{1}\left\| \textbf{u}_{\textbf{x}} \right\|_1 + \lambda_{2}\left\| \textbf{u}_t \right\|_1 + \lambda_{3}\left\| L_{\textbf{x}} \right\|_1 + \lambda_{4}\left\| \left\langle L_{\textbf{x}},\delta_t\textbf{u} \right\rangle + L_t \right\|_1 + \\ \vspace{-1.5mm}
    & \lambda_{5}h_{C}(L - L(t_p)) \text{d}t\text{d}\textbf{x} \,+ \int_{\Omega}^{} \sum_{i=2}^{\left| P(\textbf{x}) \right|} \left\| L(t_i) - L(t_{i-1}) - C p_i \right\|_1 \text{d}\textbf{x}
\end{split}
\label{eq:bardow}
\vspace{-1.5mm}
\end{equation}
The regularization terms are multiplied by $\lambda_{1}$, $\lambda_{2}$, $\lambda_{3}$, and $\lambda_{5}$, also the data term is multiplied by $\lambda_4$ (the OFC). The first three terms represent the variation of the optical flow and intensity, which are assumed to be smooth over the spatial neighborhood of each pixel $\Omega$ and over time $T$. The fifth term accounts for regions where no events are triggered, and adds up the difference between consecutive log intensity values $h_{C}(L(\textbf{x},t))$ if this difference is greater than the threshold $C$. Finally, the sixth term is the data term for regions with events, and it accounts for significant differences (greater than $\left| C p_{i} \right|$, with $p_{i} \in \{-1,1\}$ being the polarity for event $i$ in $L$. The authors mention that the work was still preliminary and needed further development to achieve accurate results. 

\textit{Filter-banks}. A differentiating approach was presented in \cite{barranco2015bio}, which proposed an OFC adapted to use phase gradient instead of intensity gradient by reconstructing the original intensity signal using the counts of events per pixel. Authors argue that this approach leads to a more stable estimation and to better deal with high-frequency textures. The model first separates textured edges and object contours and uses the local phase to estimate motion for the first case, employing the approach in \cite{barranco2014contour} to estimate motion for the object contours. Both are innovative approaches but are limited to dealing with sequences with general motion and sharp edges.
One of the problems with applying conventional optical flow methods to event processing is that event input is sparse and sometimes, active events are not enough to even form complete long edges. In \cite{brebion2021real}, authors proposed a way of generating dense inputs from the sparse flow of events. They describe an approach that generates a grid-like representation based on the inverse exponential distance surface for each event, taking into account local neighborhoods. Then, they used a filter-bank-based algorithm to estimate dense optical flow.

\textit{Contrast maximization}. Recently in \cite{shiba_2022}, authors proposed a valuable multi-scale contrast-maximization model that reaches state-of-the-art performance estimating dense optical flow. In contrast-maximization methods, the objective is to recover the sharpness of edges from the projection of the flow of events on the image plane. The approach employs a multi-reference loss function to mitigate overfitting and a time-aware estimation technique to handle occlusions. Furthermore, the multi-scale strategy improves accuracy by avoiding local minima in the optimization. The experiments show that this approach is state-of-the-art for method-based and unsupervised approaches and, that is on par with learning-based supervised networks. 

\textbf{ANN-based approaches}. 
ANNs are commonly used for event-based optical flow estimation, concretely using encoder-decoder architectures that learn hierarchical representations of motion from spatio-temporal relationships. The encoder compresses input patterns into a ``feature vector'' through convolutional or recurrent layers, while the decoder generates an output with the same size and shape as the input through deconvolutional or recurrent layers. 

\textit{Supervised}. These approaches require accurate ground truth, which is easier to achieve with simulators. Ground truth for real data requires additional hardware such as motion capture systems, that needs to be accurately synchronized with event sensors that have very high temporal resolution. Regarding the approaches, E-RAFT \citep{Gehrig_2021} is a state-of-the-art optical flow estimation model based on RAFT \citep{Teed_2020} since 2022. It uses two encoders to extract features from two consecutive event streams, builds an embedding, and estimates all-pairs correlations to iteratively produce optical flow. Recurrency is done via a GRU and a parallel context encoder that is fed into the next GRU using an additional module that warps the motion estimation from the previous temporal volumes into the next one.

Another interesting supervised learning approach for optical flow and depth estimation is EAGAN \cite{Lin_2022}. It is a generative adversarial architecture that uses a U-Net architecture with an attention-based transformer to learn temporal relationships between event sequences. EAGAN involves training both the generator and the discriminator simultaneously, using a loss function that is a combination of a supervised method for the discriminator, which requires the ground truth and the output optical flow from the generator, and a loss function similar to that of E-RAFT.

\textit{Self-Supervised}. These are generally hybrid approaches: optical flow is estimated from event information while the supervision is done with grayscale images that are either simulated or captured using hybrid sensors such as DAVIS or stereo setups (conventional camera and event sensor). The loss function is a conventional photometric loss built by warping images using event-based optical flow estimation. However, bear in mind that the low temporal frequency of conventional cameras makes them more prone to blurring or light artifacts. EV-FlowNet \citep{Zhu_2018} is an encoder-decoder inspired by U-Net and also, a baseline for optical flow estimation. Its loss function is the sum of the photometric loss and the smoothness loss that minimizes the optical flow variation. 3D-FlowNet \citep{Sun_2022} follows the same architecture but differs in that it uses 3D convolution layers to preserve temporal information. Learning requires both images and events, using a loss function similar to that of \citep{Zhu_2018}.

\textit{Unsupervised}. These approaches are purely event-based and with a loss function that usually seeks the event alignment, for example, using motion compensation. The authors of EV-FlowNet \cite{Zhu_2018}, presented an approach using only events in \cite{Zhu_new_2018}. In this work, the baseline architecture is used for the optical flow estimation, and the extended architecture with the pose model layers is used for depth and camera motion estimation. Authors use a loss function that relies on the sharpness of an image of warped events after the distortion of the event flow and its projection onto the image plane. A similar approach was followed in successive refinements of this work in \cite{hagenaars2021self, paredes2021back}.  

\begin{table}[!tbp]
     \centering

    \resizebox{0.9\textwidth}{!}{
    \begin{tabular}{ c | c | c | c | c | c | c | c | c }
        \toprule
            \textit{Type} &
            \textit{Subtype} &
            \textit{Method} & 
            \textit{PE-based} &  
            \textit{DVS Resolution} &  
            \textit{Dataset} & 
            \textit{OF Estimation} & 
            \textit{Real Time} & 
            \textit{AEE}\\
        \midrule

            \multirow{6}{*}[-2\dimexpr \aboverulesep + 2\belowrulesep + \cmidrulewidth]{\textit{TM}} &
            \multirow{2}{*}[-0.5\dimexpr \aboverulesep + \belowrulesep + \cmidrulewidth]{\textit{Local plane-fitting}} &
        \textit{\cite{Benosman_2014}} & 
            \textit{\cmark} & 
            \textit{346x260} & 
            \textit{NA} & 
            \textit{Sparse} & 
            \textit{Yes} & 
            \textit{NA}\\
        \cmidrule{3-9}

            &
            &
        \textit{\cite{Akolkar_2018}} & 
            \textit{\cmark} & 
            \textit{346x260} & 
            \textit{MVSEC} & 
            \textit{Sparse} & 
            \textit{Yes} & 
            \textit{1.59*}\\
        \cmidrule{2-9}

            &
            \textit{Variational} &
        \textit{\cite{Bardow_2016}} & 
            \textit{\cmark} & 
            \textit{128x128} & 
            \textit{NA} & 
            \textit{Dense} & 
            \textit{No} & 
            \textit{NA}\\
        \cmidrule{2-9}

            &
            \multirow{2}{*}[-0.5\dimexpr \aboverulesep + \belowrulesep + \cmidrulewidth]{\textit{Filter-banks}} &
        \textit{\cite{barranco2015bio}} & 
            \textit{\cmark} & 
            \textit{NA} & 
            \textit{NA} & 
            \textit{Sparse} & 
            \textit{Yes} & 
            \textit{NA}\\

        \cmidrule{3-9}

            &
            &
        \textit{\cite{brebion2021real}} & 
            \textit{\cmark} & 
            \textit{NA} & 
            \textit{MVSEC} & 
            \textit{Dense} & 
            \textit{Yes} & 
            \textit{NA}\\

        \cmidrule{2-9}

                &
            \textit{Contrast Maximization} &
        \textit{\cite{shiba_2022}} & 
            \textit{\cmark} & 
            \textit{NA} & 
            \textit{MVSEC, DSEC} & 
            \textit{Dense} & 
            \textit{Yes} & 
            \textit{0.42*}\\

        \midrule


            \multirow{5}{*}[-2\dimexpr \aboverulesep + \belowrulesep + \cmidrulewidth]{\textit{ANN}} &
            \multirow{2}{*}[-0.5\dimexpr \aboverulesep + \belowrulesep + \cmidrulewidth]{\textit{Supervised}} &
        \textit{\cite{Gehrig_2021}} & 
            \textit{\cmark} & 
            \textit{346x260} & 
            \textit{MVSEC, DSEC} & 
            \textit{Dense} & 
            \textit{No} & 
            \textit{NA}\\

        \cmidrule{3-9}

            &
            &
        \textit{\cite{Lin_2022}} & 
            \textit{\cmark} & 
            \textit{346x260} & 
            \textit{MVSEC} & 
            \textit{Dense} & 
            \textit{Yes} & 
            \textit{0.212}\\
        \cmidrule{2-9}

            &
            \multirow{2}{*}[-0.5\dimexpr \aboverulesep + \belowrulesep + \cmidrulewidth]{\textit{Self-Supervised}} &
        \textit{\cite{Zhu_2018}} & 
            \textit{\xmark} & 
            \textit{346x260} & 
            \textit{MVSEC} & 
            \textit{Dense} & 
            \textit{No} & 
            \textit{0.49*}\\

        \cmidrule{3-9}

            &
            &
        \textit{\cite{Sun_2022}} & 
            \textit{\cmark} & 
            \textit{256x256} & 
            \textit{MVSEC} & 
            \textit{Dense} & 
            \textit{No} & 
            \textit{0.7*}\\
        \cmidrule{2-9}

            &
            \textit{Unsupervised} &
        \textit{\cite{Zhu_new_2018}} & 
            \textit{\xmark} & 
            \textit{346x260} & 
            \textit{MVSEC} & 
            \textit{Sparse} & 
            \textit{No} & 
            \textit{0.32*}\\
        
        \midrule


            \multirow{5}{*}[-2\dimexpr \aboverulesep + 1.5\belowrulesep + \cmidrulewidth]{\textit{SNN}} &
            \multirow{3}{*}[-\dimexpr \aboverulesep + 2\belowrulesep + \cmidrulewidth]{\textit{Self-Supervised}} &
        \textit{\cite{Lee_2020}} & 
            \textit{\cmark} & 
            \textit{346x260} & 
            \textit{MVSEC} & 
            \textit{Dense} & 
            \textit{NA} & 
            \textit{1.09*}\\

        \cmidrule{3-9}

            &
            &
            
        \textit{\cite{hagenaars2021self}} & 
            \textit{\cmark} & 
            \textit{NA} & 
            \textit{MVSEC} & 
            \textit{Dense} & 
            \textit{NA} & 
            \textit{0.45*}\\
        
        \cmidrule{3-9}
        
            &
            &
        \textit{\cite{kosta_2022}} & 
            \textit{\cmark} &
            \textit{NA} & 
            \textit{MVSEC, DSEC} & 
            \textit{Dense} & 
            \textit{NA} & 
            \textit{0.44*}\\
            
        \cmidrule{2-9}

            &
            \multirow{2}{*}[-0.5\dimexpr \aboverulesep + 2\belowrulesep + \cmidrulewidth]{\textit{Unsupervised}} &
        \textit{\cite{Haessing_2018}} & 
            \textit{\cmark} & 
            \textit{304x240} & 
            \textit{NA} & 
            \textit{Dense} & 
            \textit{Yes} & 
            \textit{11\%}\\
        \cmidrule{3-9}

            &
            &
        \textit{\cite{Lee_2021}} & 
            \textit{\xmark} & 
            \textit{256x256} & 
            \textit{MVSEC} & 
            \textit{Dense} & 
            \textit{NA} & 
            \textit{0.56*}\\

        \hline  
        \bottomrule            
    \end{tabular}  
    }
    \caption{Most relevant optical flow algorithms using event cameras. PE refers to ``Purely Event'' based, NA refers to ``Not Available'', and AEE stands for ``Average End-point Error''. Regarding the results, the reported AAE is generally the average error for all the testing sequences in the dataset. Otherwise, $*$ indicates that the reported AEE is for a particular sequence proposed in the work.}
    \label{tab:OF_methods}
    \vspace{-4.5mm}
\end{table}
\textbf{SNN-based approaches}. SNN-based learning models have not reached their full potential yet and thus, they are not on par in accuracy with deep learning models. 

\textit{Self-Supervised}. In recent research, the integration of SNNs and ANNs has been explored to overcome the lack of backpropagation in SNNs. One such implementation is the Spike-FlowNet model, which combines SNN and ANN layers to create an encoder-decoder architecture. The encoder uses SNN layers, while the decoder is made up of ANN layers. This hybrid approach has been shown to save up to 17.6\% of energy. The authors of \cite{Lee_2020} trained the ANN using the sum of two loss functions inspired by \cite{Zhu_2018}. These are the photometric loss, which minimizes differences between two grayscale images, and the smoothness loss, which acts as a normalization term. The output is then passed into the SNN block that uses the surrogate gradient method \cite{Werbos_1990} for training. In \cite{hagenaars2021self}, authors used two different RNNs based on EV-FlowNet and FireNet architectures, along with various SNN alternatives for optical flow estimation. This work achieved good results in terms of accuracy but has high computational complexity. Recently, Adaptive-SpikeNet \cite{kosta_2022} aimed to adapt successful ANN models to fully-spiking models using SNNs. It combined the surrogate gradient method with self-supervised learning. The adapted models showed better efficiency than previous ANN and hybrid approaches, although ANNs still achieved lower error. In terms of energy consumption, SNN models were 5x more efficient.

\textit{Unsupervised}. One of the first implementations in an embedded system was proposed in \cite{Haessing_2018}, with an SNN based on an adaptation of the Barlow \& Levick model \cite{Barlow_1965}, using Direction Sensitive neurons. The network was implemented in the IBM TrueNorth Neurosynaptic System, achieving real-time operation with an energy consumption of $80~mW$. Fusion-FlowNet was presented in \cite{Lee_2021} and it combines event-based and frame-based cameras to compute dense optical flow, in a model based on U-Net \cite{Ronneberger_2015}. The encoder uses SNN layers for the events and ANN convolutional layers for the frames, the residual blocks, and the decoder. SNN layers are endowed with signed activities, and learning is enabled via the surrogate gradient method to enable back-propagation. 

\textbf{Datasets}. Collecting datasets is essential for learning-based methods but, it is also a very time-consuming task that requires ensuring diversity, completeness, high reliability, and accuracy. 

\textit{MVSEC} \cite{Zhu_2018_dataset} builds on previous attempts \cite{Barranco_2016_dataset, Binas_2017_dataset} that were limited due to their size or lack of diversity. The dataset was built using a stereo camera and an IMU, a pair of DAVIS neuromorphic sensors with integrated IMUs, a LiDAR sensor that provides ground truth, and a GPS unit. It contains 14 indoor and outdoor sequences including day and night scenes, relevant to assess the high-dynamic range of the neuromorphic sensors. 

\textit{DSEC} overcomes some of the limitations in MVSEC \cite{ghosh2022multi} as scenes that lack some ground truth or source data, or poor 3D data for some scenes due to their stereo-rig baseline distance compared to the maximum depth of these scenes. In this case, the authors use two Prophesee Gen3.1 neuromorphic sensors, a stereo pair of cameras, a LiDAR, and a GPS sensor. All 53 sequences in this dataset were collected from a car, driving in urban, suburban, and rural areas, including day and night sequences, with some challenging artifacts such as a car driving out of a tunnel. Table \ref{tab:eventdatasets} summarizes the most relevant features of both datasets. 

\subsubsection{Ego-Motion}
\label{sec:egomotion}
The biggest challenge for autonomous robotic systems is to execute fast and secure trajectories while navigating. To do it, their relative pose with respect to the scene needs to be accurately estimated. For the last few years, precise navigation solutions \cite{Romero_2022, foehn2021time} have demonstrated that current control algorithms for aggressive maneuvers are robust enough. However, all these precise solutions are not computed from the robot perception subsystem estimating their own 3D self-motion or ego-motion. Most of these solutions only work on controlled scenarios and use motion-capture (MoCap) systems, which have been widely proven to be much faster and more reliable than any other solution for real-time pose estimation. However, these are expensive systems that need active cameras and can only be used indoors. 
 
The set of methods that allow estimating the robot motion (translation and rotation) relative to a reference frame is known as visual odometry (VO) or ego-motion. Efficient onboard solutions with monocular cameras, stereo cameras, or laser \cite{Jeon_2018, Liu_2021} have been proven sufficient for motion estimation. However, their latency still limits the processing pipeline in dynamic environments or at high speeds. Brain-inspired solutions are good candidates for onboard processing due to their low computational complexity and energy consumption. Also, these solutions leverage the agility of the robot by minimizing the latency of perception which is a crucial constraint to achieving real-time navigation at high speeds \cite{falanga2020dynamic}. Notice that some methods in Section \ref{subsubsec:OF} also address the ego-motion estimation problem usually as a by-product when estimating optical flow. To avoid redundancy, this section focuses only on ego-motion estimation as shown in Table \ref{tab:egomotion_methods}. The presented works are divided into monocular and stereo approaches, and categorized attending to the event stream data representation. 
\begin{table}[!tbp]
\newcommand{\tabincell}[2]{\begin{tabular}{@{}#1@{}}#2\end{tabular}}
    \centering

    \resizebox{0.9\textwidth}{!}{
    \begin{tabular}{c|c|c|c|c|c|c}
        \toprule
        \textit{Event Representation} & \textit{Method} & \textit{PE-based} &  \textit{Sensor} & \textit{Resolution} & \textit{Dataset}  & \textit{Accuracy}\\
        
        \cmidrule{1-7}
        
        \multirow{6}{*}[-4em]{Event Accumulation} 
        
        & \tabincell{c}{\textit{\cite{Mueggler_2014}}} & \textit{\cmark} & \tabincell{c}{\textit{Monocular}} & \tabincell{c}{\textit{128 x 128}} & \tabincell{c}{\textit{Custom}} & \tabincell{c}{\textit{Trans. error: }$0.108\pm 0.078$ m\\\textit{Orient. error: } $5.1\pm 2.4$º}\\
        \cmidrule{2-7}

        & \tabincell{c}{\textit{\cite{Kim_2016}}} & \textit{\cmark} & \tabincell{c}{\textit{Monocular}}  &  \tabincell{c}{\textit{NA}} & \tabincell{c}{\textit{Custom}}& \tabincell{c}{\textit{Qualitative} \\\textit{Results Only}}\\ 
        \cmidrule{2-7}

        & \tabincell{c}{\textit{\cite{Rebecq_2017}}} & \textit{\cmark} & \tabincell{c}{\textit{Monocular}}  &  \tabincell{c}{\textit{240 x 180}} & \tabincell{c}{\textit{Custom}}& \tabincell{c}{\textit{ Trans. error: }$\leq0.2\%$ \\\textit{Orient. error: } $\leq3$º}\\
        \cmidrule{2-7}

        & \tabincell{c}{\textit{\cite{Vidal_2017}}} & \textit{\xmark} & \tabincell{c}{\textit{Monocular}\\\textit{Events + Frames + IMU}}  &  \tabincell{c}{\textit{240 x 180}} & \tabincell{c}{\textit{Event Camera Dataset}}&  \tabincell{c}{\textit{Trans. error: } $\leq 0.24\%$ \\\textit{Orient. error: } $\leq 0.06$º}\\ 
        \cmidrule{2-7}

        & \tabincell{c}{\textit{\cite{Nguyen_2019}}} & \textit{\cmark} & \tabincell{c}{\textit{Monocular}}  &  \tabincell{c}{\textit{240 × 180}} & \tabincell{c}{\textit{Event Camera Dataset}}& \tabincell{c}{\textit{ Trans. error: }$0.036$ m \\\textit{Orient. error: } $2.341$º}\\
        \cmidrule{2-7}
        
        & \tabincell{c}{\textit{\cite{Colonnier_2021}}} & \textit{\cmark} & \tabincell{c}{\textit{Monocular}}  &  \tabincell{c}{\textit{240 x 180}} & \tabincell{c}{\textit{Custom}}& \tabincell{c}{\textit{Trans. error: }$<5\%$ \\\textit{Orient. error: } $<4$º}\\
        \midrule

        \multirow{2}{*}[-0.8em]{Probabilistic} 
        & \tabincell{c}{\textit{\cite{Censi_2014}}} & \tabincell{c}{\textit{\xmark}} & \tabincell{c}{\textit{Monocular}\\\textit{Events + Frames}}  &  \tabincell{c}{\textit{128 x 128}} & \tabincell{c}{\textit{Custom}} & \tabincell{c}{\textit{Trans. error: } \textit{NA} \\\textit{Orient. error: } $\leq 1$º}\\ 
        \cmidrule{2-7}
        
        & \tabincell{c}{\textit{\cite{Gallego_2018}}} & \textit{\xmark} & \tabincell{c}{\textit{Monocular}\\\textit{Events + RGBD Map}}  &  \tabincell{c}{\textit{128 x 128}} & \tabincell{c}{\textit{Custom}}& \tabincell{c}{\textit{Trans. error: } $\leq 1.8\%$ \\\textit{Orient. error: } $\leq 1.04$º}\\ 
        \midrule

        \multirow{1}{*}{Events Volume} 
        & \tabincell{c}{\textit{\cite{Zhu_2017}}} & \textit{\xmark} & \tabincell{c}{\textit{Monocular}\\\textit{Events + IMU}}  &  \tabincell{c}{\textit{240 x 180}} & \tabincell{c}{\textit{Event Camera Dataset}}& \tabincell{c}{\textit{Trans. error: } $\leq 2.57\%$ \\\textit{Orient. error: } $\leq 0.27$º/m}\\ 
        \midrule

        \multirow{3}{*}[-1.6em]{Time Surfaces}               
        & \tabincell{c}{\textit{\cite{ZhouYi_2021}}} & \textit{\cmark} & \tabincell{c}{\textit{Stereo}}  &  \tabincell{c}{\textit{346 x 260}} & \tabincell{c}{\textit{MVSEC *} \\ \textit{Event Camera Dataset *}}& \tabincell{c}{\textit{Trans. error: } $\leq 0.053$ m/s \\\textit{Orient. error: } $\leq 1.93$º/s}\\ 
        \cmidrule{2-7}
        & \tabincell{c}{\textit{\cite{Hadviger_2021}}} & \textit{\cmark} & \tabincell{c}{\textit{Stereo}}  &  \tabincell{c}{\textit{346 x 260}} & \tabincell{c}{\textit{MVSEC *} \\ \textit{DSEC *}}& \tabincell{c}{\textit{Trans. error: }$\leq 8.76\%$ \\\textit{Orient. error: } $\leq 1.17$º/m}\\ 
        \cmidrule{2-7}
        
        & \tabincell{c}{\textit{\cite{Zuo_2022_devo}}} & \textit{\cmark} & \tabincell{c}{\textit{Stereo}\\\textit{Events + RGBD}}  &  \tabincell{c}{\textit{640×480}} & \tabincell{c}{\textit{Custom}}&\tabincell{c}{\textit{Trans. error: } $\leq 0.0142$m/s \\\textit{Orient. error: } $\leq 1.57$º/s}\\ 
        \midrule

        \multirow{1}{*}{ANN Unsupervised} 
        & \tabincell{c}{\textit{\cite{ye2020unsupervised}}} & \textit{\cmark} & \tabincell{c}{\textit{Monocular}}  &  \tabincell{c}{\textit{NA}} & \tabincell{c}{\textit{MVSEC *}}& \tabincell{c}{\textit{ Trans. error: }$3.65$º \\\textit{Orient. error: } $0.23$º}\\   
        
        \hline       
        \bottomrule            
    \end{tabular}  
    }
    \caption{Most relevant ego-motion algorithms using event cameras. PE refers to ``Purely Event'' based, and NA stands for ``Not Available''. The accuracy values are given as the mean of the individually presented results of every sequence from the dataset used. * stands for not using all the sequences of the dataset.}
    \label{tab:egomotion_methods}
\vspace{-1.5mm}
\end{table}

\textbf{Monocular Ego-Motion}. Most pose estimation methods rely on a single camera, avoiding the need to solve the data association problem of multi-view camera systems. Monocular approaches have been shown to be sufficient to estimate the ego-motion.

\textit{Event Accumulation}. The first onboard low-latency approach for 6 degrees of freedom estimation for aggressive maneuvers using an event camera was presented in \cite{Mueggler_2014}. In this approach, 
lines are built from sparse events and for each new event, the method decides which close-by lines are to be updated. The pose estimation is computed by minimizing the sum of squared distances between the new events and the reprojection of each line, that belongs to a known pattern. The method estimates the pose under very challenging conditions but requires this pattern. 

In \cite{Kim_2016}, the authors describe a real-time joint estimation of 6-DoF camera motion, 3D structure, and intensity using three decoupled EKF-based probabilistic filters. The proposed approach shows good results in unstructured environments with no prior knowledge of the scene. EVO \cite{Rebecq_2017} is a purely geometric event-based tracking and mapping method that runs on the CPU in real time and outputs up to several hundred pose estimates per second. It involves running a 3D reconstruction algorithm \cite{Rebecq_2018}, in parallel with an event-based image-to-model alignment for tracking. In frame-based direct methods, this alignment between two frames is computed as a minimization of its photometric error \cite{Forster_2014}. However, since event cameras do only respond to edges in the scene, this photometric error is replaced by a geometric alignment between two edge images as shown in Eq. \ref{eq:rebecq_2017}. 
\vspace{-1.5mm}
\begin{equation}
\underset{\Delta P}{arg\,min} \sum_{\textbf{x}}(M(\mathcal{W}(\textbf{x},\Delta P))-\Gamma(\mathcal{W}(\textbf{x},P)))^{2}
\label{eq:rebecq_2017}
\vspace{-1.5mm}
\end{equation}
That is indeed the inverse compositional Lucas-Kanade alignment algorithm \cite{baker2004lucas}, where $M$ corresponds to the \textit{template} of the projected semi-dense map that is incrementally built, which is used to compare with the current event accumulated image $\Gamma$. To achieve this, a wrap-rigid-body transformation $\mathcal{W}$, at a point $\textbf{x}$ in the image plane of the template $M$, is minimized to compute the pose estimate $P$ (Eq. \ref{eq:rebecq_2017}). It is a minimization for the incremental pose estimation $\Delta P$ that is, the complete set of transformations of the map from the origin to the current pose. Since transformations are relative, this method is prone to incremental error accumulation. 

UltimateSLAM is a holistic approach \cite{Vidal_2017} which consists of a pipeline for robust and accurate state estimation that tightly combines events, frames, and inertial measurements. The main idea is to build event frames by aggregating spatiotemporal windows of events, and then use the FAST corner detector to extract features that will be tracked using the Lucas-Kanade tracker \cite{LKTracker}. The camera pose is updated fusing conventional frames and virtual motion-compensated event frames \cite{Rebecq_2017_vio}, with a final refinement fusing inertial cues.  
This final problem is solved as a joint optimization whose cost function $J$ shown in Eq. \ref{eq:vidal_2017}  consists of two reprojection errors $e^{i}$ of the observations from the event-based ($i=0$) and the frame-based ($i=1$) cameras, plus the inertial error $e_{s}$.
\vspace{-1.5mm}
\begin{equation}
J = \sum_{i=0}^{1}\sum_{k=1}^{K}\sum_{j \in \mathcal{J}(i,k)} e^{i,j,k^{T}} W_{r}^{i,j,k} e^{i,j,k} + \sum_{k=1}^{K-1}e_{s}^{k^{T}}W_{s}^{k}e_{s}^{k}
\label{eq:vidal_2017}
\vspace{-1.5mm}
\end{equation}
In Eq. \ref{eq:vidal_2017}, $k$ is the frame index, $j$ the landmark index, and $\mathcal{J}(i,k)$ is the set of landmark indices for each sensor $i$ and frame $k$. The terms $W_{r}^{i,j,k}$ and $W_{s}^{k}$ are the information matrices of the landmarks and inertial measurements respectively. Their results show the complementarity of conventional frames and events, boosting accuracy in $130\%$ compared to using only events and IMU data.

\cite{Nguyen_2019} proposed a Stacked Spatial LSTM Network (SP-LSTM) to predict the camera pose by taking just event-wrapped images as input. In this work, features from the event images are extracted using a CNN. The output of the network is reshaped and used as input for the SP-LSTM module. This network aims to learn spatial dependencies in the feature space, which can improve the relocalization of the camera pose. This network has an inference time of $5 ms$, which can be considered a real-time system but is not enough for high-speed robotic applications. A 6-DoF implementation relying on an EKF for pose estimation was proposed in \cite{Colonnier_2021}. The key of the algorithm is the proposed event-to-map association. First, a 3D point cloud is initialized by event projection to the parallel image plane at a fixed distance, assuming that the sensor is initially static. Then, a pixel-to-3D-point association is performed and subsequently tracked for pose estimation with an EKF, continuously updating the map to track the event-to-map association. The main advantage of this event-by-event method is that it updates the overall estimation with every new incoming event and thus, it greatly reduces latency.

\textit{Probabilistic}. \cite{Censi_2014} solves visual odometry using an event sensor and a conventional camera. The solution is limited, and only rotations are accurately computed while translation estimation is very noisy. A differential approximation is described in \cite{Gallego_2018}, in which authors tackle full-motion estimation for augmented/virtual reality (AR/VR) purposes, using a single event camera, and a photometric depth map built with a traditional dense 3D reconstruction and conventional images. Although the estimation is accurate, the approach requires RGB-D sensors or an additional method to estimate depth from conventional frames. 

\textit{Event Volume}. EVIO \cite{Zhu_2017} represents the first event-based feature tracker algorithm to fuse event data streams with an inertial measurement unit. Features are detected within the orthogonal projection of the events onto the image plane, using the FAST corner detector \cite{Rosten2010FasterAB}. Finally, it fuses IMU data with EKF to improve the tracking accuracy. 

\textit{ANN Unsupervised}. The authors of \cite{ye2020unsupervised}, presented an approach purely based on events with an Evenly-Cascaded convolutional Network (ECN), using an encoder-decoder architecture to estimate the depth and a second neural network that takes consecutive slices of events to estimate the camera pose. Then, these two neural networks are used to estimate the optical flow. 

\textbf{Stereo Ego-Motion}. There are also stereo implementations that try to mimic the dual-eye capabilities of biological systems and exploit stereo advantages for depth estimation (a very detailed survey on stereo event-based vision can be found in \cite{Steffen_2019}). 

\textit{Time Surfaces}. ESVO \cite{ZhouYi_2021} is the first parallel tracking-and-mapping approach using a stereo event camera. The mapping module builds a semidense 3D map of the scene by fusing depth estimates from multiple viewpoints (obtained by spatio-temporal consistency) in a probabilistic fashion. Tracking of event features is performed by fusing the map and the stream of events. Event features are extracted from spatio-temporal surfaces also known as time surfaces $\mathcal{S}(\textbf{x},t) = e^{ -(t-t_{last})/\tau}$ \cite{Lagorce_2017} which encode the motion history of the edges. The time surface $\mathcal{S}$ is a 2D map that is updated according to the timestamp of the last event $t_{last}$ occurring at a pixel location $\textbf{x}$ and a decay rate parameter $\tau$. Looking at time surfaces, large values show the locations where the current edges are. The estimation is done by aligning time surfaces over time using the depth maps, as if time surfaces were artificial images. In fact, the tracking method consists of aligning these regions with the computed depth map once it is warped to the time surface frame by a candidate pose. Thus, to solve the compositional Lucas-Kanade method, authors compute the motion of the camera as an optimization with the cost function shown in Eq. \ref{eq:esvo_lk}.
\vspace{-1.5mm}
\begin{equation}
F(\Delta \theta) = \sum_{\textbf{x}\in S^{F_{ref}}}(\mathcal{S} (\mathcal{W}(\mathcal{W}(\textbf{x},\delta ;\Delta\theta);\theta),k))^2
\label{eq:esvo_lk}
\vspace{-1.5mm}
\end{equation}
Where $\theta$ stands for the motion (orientation and translation),  $S^{F_{ref}}$ is the set of pixels  $\textbf{x}$ in the reference frame $F_{ref}$ with a valid inverse depth $z$, and $\mathcal{W}$ is a wrapping function which aligns the semi-dense map with the current time surface at time $k$. 

More recently, \cite{Hadviger_2021} proposed an event stereo visual odometry method that relies on features that are matched through consecutive left and right event time surfaces. The method does not produce pose information at a fixed frequency, but dynamically, and it only depends on the number of incoming events to ensure sufficient visual information for reliable feature detection. Finally, \cite{Zuo_2022_devo} presented a hybrid stereo setup with a depth camera and an event camera. This approach first extracts edges from time surface maps then assigns depth to them using an RGB-D camera, and finally registers subsequent views by 3D-2D edge alignment. Their results outperform ESVO and are robust under challenging illumination conditions. However, the requirement of the depth camera makes it biologically unplausible. 

\textbf{Datasets}. A first attempt for a small dataset was done in \cite{Barranco_2016_dataset}, including several sequences of different general motions on unstructured scenes. The dataset contains optical flow, depth maps, and 3D motion ground truth data. As for large-scale datasets, the main efforts have been \textit{MVSEC} and \textit{DSEC} datasets, already described in Section \ref{subsubsec:OF}. 

\begin{table}[!t]
    \centering
\resizebox{0.9\textwidth}{!}{%
    \begin{tabular}{llccclll}
    \toprule
    \multicolumn{1}{c|}{\textit{Dataset}} & \multicolumn{1}{c|}{\textit{Sensor}} & \multicolumn{1}{c|}{\textit{Resolution}} & \multicolumn{1}{c|}{\textit{\begin{tabular}[c]{@{}c@{}}Baseline\\ (cm)\end{tabular}}} & \multicolumn{1}{c|}{\textit{\begin{tabular}[c]{@{}c@{}}FOV\\ (º)\end{tabular}}} & \multicolumn{1}{c|}{\textit{Illumination}} & \multicolumn{1}{c|}{\textit{\begin{tabular}[c]{@{}c@{}}Max. depth\\ (m)\end{tabular}}} & \multicolumn{1}{c}{\textit{Total}} \\ 
    \midrule
    
    \multicolumn{1}{l|}{\textit{\begin{tabular}[c]{@{}l@{}}MVSEC\\ \cite{Zhu_2018_dataset}\end{tabular}}}
    & \multicolumn{1}{l|}{\textit{Davis346}} & \multicolumn{1}{c|}{\textit{346x260}} & \multicolumn{1}{c|}{\textit{10}} & \multicolumn{1}{c|}{\textit{74.8}} & \multicolumn{1}{l|}{\textit{\begin{tabular}[c]{@{}l@{}}Indoor/outdoor\\ Day/night\end{tabular}}} & \multicolumn{1}{c|}{\textit{(Indoor) 8.4}} & \textit{\begin{tabular}[c]{@{}l@{}}14 sequences\\ (4083 s)\end{tabular}} \\
    \midrule
    
    \multicolumn{1}{l|}{\textit{\begin{tabular}[c]{@{}l@{}}DSEC\\ \cite{Gehrig_Dataset_2021}\end{tabular}}}& \multicolumn{1}{l|}{\textit{\begin{tabular}[c]{@{}l@{}}2 x Prophesee \\ Gen3\end{tabular}}} & \multicolumn{1}{c|}{\textit{640x480}} & \multicolumn{1}{c|}{\textit{60}} & \multicolumn{1}{c|}{\textit{60.1}} & \multicolumn{1}{l|}{\textit{\begin{tabular}[c]{@{}l@{}}Outdoor\\ Day/night\end{tabular}}} & \multicolumn{1}{c|}{\textit{50}} & \textit{\begin{tabular}[c]{@{}l@{}}53 sequences\\ (3193 s)\end{tabular}} \\
    \midrule    
    \multicolumn{1}{l|}{\textit{\begin{tabular}[c]{@{}l@{}}TUM-VIE\\ \cite{klenk2021tum}\end{tabular}}} & \multicolumn{1}{l|}{\textit{2 x Prophesee Gen4}} & \multicolumn{1}{c|}{\textit{1280x720}} & \multicolumn{1}{c|}{\textit{11.84}} & \multicolumn{1}{c|}{\textit{90}} & \multicolumn{1}{l|}{\textit{\begin{tabular}[c]{@{}l@{}}Indoor/outdoor\\ Artifacts, day/night\end{tabular}}} & \multicolumn{1}{c|}{\textit{NA}} & \textit{\begin{tabular}[c]{@{}l@{}}21 sequences\\ (2880 s)\end{tabular}} \\ \midrule
    
    \multicolumn{1}{l|}{\textit{\begin{tabular}[c]{@{}l@{}}VECtor\\ \cite{Gao_2022}\end{tabular}}} & \multicolumn{1}{l|}{\textit{2 x Prophesee Gen3}} & \multicolumn{1}{c|}{\textit{640x480}} & \multicolumn{1}{c|}{\textit{17}} & \multicolumn{1}{c|}{\textit{82}} & \multicolumn{1}{l|}{\textit{\begin{tabular}[c]{@{}l@{}}Indoor/outdoor\\ Day/night\end{tabular}}} & \multicolumn{1}{c|}{\textit{5 (indoor) }} & \textit{\begin{tabular}[c]{@{}l@{}}18 sequences\\ ($>$620 s)\end{tabular}} \\
    
\midrule
    \multicolumn{1}{l|}{\textit{\begin{tabular}[c]{@{}l@{}}EV-IMO\\ \cite{Mitrokhin_2019}\end{tabular}}}
    & \multicolumn{1}{l|}{\textit{Davis}} & \multicolumn{1}{c|}{\textit{346x260}} & \multicolumn{1}{c|}{\textit{NA}} & \multicolumn{1}{c|}{\textit{80}} & \multicolumn{1}{l|}{\textit{\begin{tabular}[c]{@{}l@{}}Indoor\end{tabular}}} & \multicolumn{1}{c|}{\textit{NA}} &  \textit{\begin{tabular}[c]{@{}l@{}}6 sets containing \\ 19 sequences (1920 s)\end{tabular}} \\
    \midrule
    
    \multicolumn{1}{l|}{\textit{\begin{tabular}[c]{@{}l@{}}MOD and MOD++\\ \cite{Sanket_2019} and\\ \cite{Parameshwara_2020}\end{tabular}}}
    & \multicolumn{1}{l|}{\textit{None}} & \multicolumn{1}{c|}{\textit{346x260}} & \multicolumn{1}{c|}{\textit{NA}} & \multicolumn{1}{c|}{\textit{NA}} & \multicolumn{1}{l|}{\textit{\begin{tabular}[c]{@{}l@{}}Outdoor\\ Day\end{tabular}}} & \multicolumn{1}{c|}{\textit{NA}}  & \textit{\begin{tabular}[c]{@{}l@{}}6 (MOD) + 2 (MOD++) \\ sequences\end{tabular}} \\
    \midrule

    \multicolumn{1}{l|}{\textit{\begin{tabular}[c]{@{}l@{}}EV-IMO2\\ \cite{Burner_2022}\end{tabular}}}
    & \multicolumn{1}{l|}{\textit{\begin{tabular}[c]{@{}l@{}}2 x Prophesee Gen3 \\ Samsung DVS \end{tabular}}} & \multicolumn{1}{c|}{\textit{640x480}} & \multicolumn{1}{c|}{\textit{22}} & \multicolumn{1}{c|}{\textit{75}} & \multicolumn{1}{l|}{\textit{\begin{tabular}[c]{@{}l@{}}Indoor \\ low light\end{tabular}}} & \multicolumn{1}{c|}{\textit{NA}} & \textit{\begin{tabular}[c]{@{}l@{}}173 sequences \\ (2460 s)\end{tabular}} \\
    
    \hline       
    \bottomrule
    
    \end{tabular}%
}
\caption{Summary of parameters of the most popular event-based datasets. NA stands for ``Not Available''.}
\label{tab:eventdatasets}
\end{table}

\textit{TUM-VIE} \cite{klenk2021tum} contains handheld and head-mounted recordings mostly indoors. It also includes some sequences with rapid motion during sports practice with changing illumination conditions. It was recorded ensuring hardware synchronization, using a stereo pair of conventional cameras, two Prophesee Gen4 event sensors, an inertial sensor, and the MoCap system at 120 Hz.

Finally, the \textit{VECtor} dataset \cite{Gao_2022} includes two sets of sequences: 1) 12 small-scale indoor sequences, some of them recorded under low illumination conditions; 2) 6 large-scale sequences recorded while walking or driving a scooter. The setup includes a stereo pair of Prophesee Gen3, a stereo pair of conventional cameras, an RGB-D camera, an IMU, a LiDAR, a Laser Scanner, and a MoCap system that provides ground truth. All datasets are summarized in Table \ref{tab:eventdatasets}. 

\subsection{Mapping}
\label{subsec:mapping}
When deploying a robot in an unknown scenario the map might not be available. To achieve truly autonomous capabilities, the robot must be able to build this map in real-time while it moves. First, the robot estimates depth and then, it builds the map as an aggregation of sparse 3D features. This section describes depth estimation methods and common mapping representations with events.

\subsubsection{Depth Estimation and 3D Reconstruction}
Autonomous navigation in any environment requires a 3D representation of the scene to plan the trajectory to follow or simply, to avoid collisions. Depth cues are crucial for mapping or planning algorithms, especially if environments are dynamic, or in the most simple case, for reactive controllers. Experiments such as \cite{Hu_2019}, prove that edges are more relevant than color to estimating depth. Therefore, as event cameras do only respond to salient edges and textures, they are the best candidates for accurate depth estimation of the scene. In this section, we present stereo and monocular event-based solutions.

\textbf{Monocular Depth Estimation}. Geometrical-based approaches using a unique camera for 3D semi-dense depth reconstruction of the space require information about the moving camera, which means using its ego-motion as input. Thus, a direct point cloud generation cannot be achieved. 

EMVS \cite{Rebecq_2018} was the first multi-view stereo approach that reconstructed the 3D structure of the scene in real-time using a single event camera. However, in this approach, the model obtained the camera pose from an external tracking system. In \cite{Gallego_depth_2018}, the authors presented a unifying framework to simultaneously solve the estimation of 3D motion, depth, and optical flow. The method looks for point trajectories on the image plane that best fit the event data by reconstructing the scene dynamics. Additionally, the framework outputs motion-corrected event images using an objective function to measure how well the events are aligned along the candidate trajectories using contrast maximization. Finally, \cite{Gallego2019FocusIA} provides a comprehensive list of objective/focus-loss functions for motion compensation approaches. 

\cite{Hidalgo_2020} proposed the first system for dense depth monocular estimation using only events with a recurrent convolution network. This is done to leverage the temporal consistency of the event stream. The proposed methodology shows better accuracy results than existing methods. The loss function, Eq. \ref{eq:hidalgo}, is the weighted sum of the scale-invariant gradient matching loss $\mathcal{L}_{k,si}$ and the multi-scale-invariant gradient matching loss $\mathcal{L}_{k,grad}$. This last term allows the network to predict smooth depth maps, enforcing sharp depth discontinuities.
\vspace{-1.5mm}
\begin{equation}
\begin{split}
\mathcal{L}_{k,si}=\frac{1}{n}\sum_{\textbf{x}}(R_{k}(\textbf{x}))^2-\frac{1}{n^2}\left ( \sum_{\textbf{x}} R_{k}(\textbf{x})\right )^2 \\
\mathcal{L}_{k,grad}=\frac{1}{n}\sum_{s}\sum_{\textbf{x}}\left \| \nabla_{x}R_{k}^{s}(\textbf{x}) \right \|_{1}+ \left \| \nabla_{y}R_{k}^{s}(\textbf{x}) \right \|_{1}
\end{split}
\label{eq:hidalgo}
\vspace{-1.5mm}
\end{equation}
Where $R_k$ and $R_{k}^{s}$ are the residual of the depth, and that same residual at scale $s$ respectively, for $n$ valid ground-truth pixels $\textbf{x}$ (more precisely, for a sequence of ground-truth depth maps).

A learning-based monocular depth estimation was proposed by \cite{Gehrig_ram_2021} using events and frames to capture richer information from the scene. Due to the architectural limitation of traditional Recurrent Neural Networks (RNN) \cite{Siam_2017} for processing asynchronous events and synchronous frames, authors propose a Recurrent Asynchronous Multi-modal (RAM) network. This network enables the processing of any irregular data from multimodal sensors. The main drawback of this approach is the need to use both frames and event streams. Finally, in \cite{Haessig_2019}, authors presented a neuromorphic approach employing an SNN that uses depth from defocus (DFD), proposing a hardware-friendly implementation that estimates sparse depth. 

\begin{figure*}[!t]
	\centering
	\includegraphics[width=0.9\textwidth]{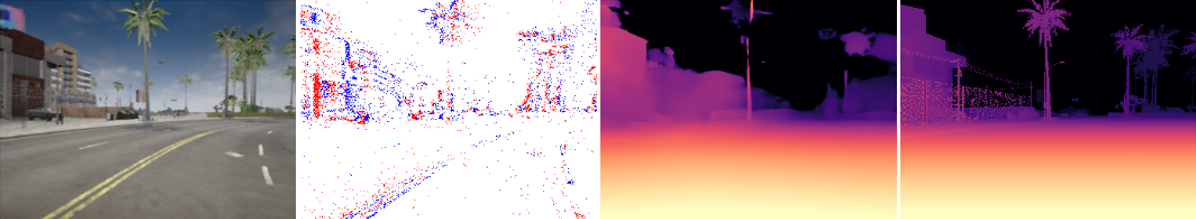}
        \vspace{-2.5mm}
	\caption{Qualitative results comparing depth prediction (extracted from \cite{Hidalgo_2020}). From left to right: frames, event stream, event-based predicted depth, and ground-truth depth.}
	\label{fig_depth_hidalgo}
        \vspace{-2.5mm}
\end{figure*}

\textbf{Stereo Depth Estimation}. Most stereo depth works to tackle the problem known as \textit{instantaneous} stereo depth estimation. This means generating a 3D point cloud in a very short time (ideally, event by event) using a stereo pair and ensuring clock synchronization. Synchronization is limited due to factors such as different pixel sensitives that can induce different activation/deactivation timestamps, or sensor bus transmission congestion that will delay the signal. The 3D reconstruction general pipeline follows a two-step process: 1) solving the stereo event matching correspondence (which is the most computationally intensive step), for example, using asynchronous epipolar constraint as shown in \cite{Benosman_2011, Rogister_2012}; 2) computing each 3D point by triangulation following classical solutions \cite{Hartley_2004}. 

\cite{Piatkowska_2014} estimated depth using a dynamic cooperative network for asynchronous stereo matching. 
The key aspect of this disparity calculation approach is that it does not only consider the temporal similarity of the events but also the spatiotemporal neighborhood. The first dense depth map reconstruction was presented in \cite{Dongqing_2017}. The work proposed a discriminative event feature descriptor, which showed invariance to changes in translation, rotation, and scale. It also proposed an event-enhancing method that is crucial for finding matches between the stereo event streams. More recently, authors in \cite{Zhou_2018_Semi} proposed a non-instantaneous 3D reconstruction algorithm with a stereo event-based configuration by matching temporal coherence of time-surface maps \cite{Lagorce_2017} from left and right event sensors. Depth estimate is solved via energy minimization exploiting the spatiotemporal consistency, instead of the classical event matching plus triangulation pipeline.

In \cite{Risi_2021}, authors proposed an instantaneous stereo depth estimation approach by using SNNs with events as input. The intention of using SNNs for the stereo-matching computation is to reduce the intrinsic bottleneck of this process. This architecture is implemented on a mixed-signal neuromorphic processor, leveraging this to a more similar brain-inspired solution. Finally, the work in \cite{ranccon2021stereospike} uses a hardware-friendly fully-spiking SNN that is fed with two trains of events from different DVS neuromorphic sensors, using an encoder-decoder architecture. Interestingly, this network achieves state-of-the-art accuracy, outperforming its image-based equivalent ANN. A thorough review of depth estimation can be found in \cite{Furmonas_2022}.

\textbf{Datasets}. There have been different attempts to provide datasets that include structured and non-structured scenarios, mostly indoors due to the limitations of RGB-D cameras that were used to collect ground truth depth. However, nowadays the most popular datasets for depth estimation are \textit{MVSEC} already described in Section \ref{subsubsec:OF}, and \textit{TUM-VIE} described in Section \ref{sec:egomotion}.

\subsubsection{Mapping Representation}
One of the most important stages in the pipeline of autonomous navigation is the representation of the environment, enabling the robot to plan its routes. Most works in the literature consider two representations: metric and topological maps. 

\textbf{Metric Maps}. They consider geometric information of the environment and are built to establish a fixed reference frame. Solutions range from sparse landmark-based and occupancy grid maps to 3D semi-dense or fully-dense reconstructions. Occupancy grids are the most popular kind of metric maps \cite{Oleynikova_2017, hornung13auro} due to their good localization accuracy vs. computational complexity trade-off. These maps are represented as an array of 2D pixels or 3D voxels that encode occupancy probability.

\textbf{Topological Maps}. Instead of considering the geometric relationship between landmarks, these maps gather adjacency information. Their most common representation are graphs where each node encodes a place or region of interest, and links that represent the adjacency information of the connection between nodes \cite{Blchliger2018TopomapTM}. This representation reduces the computational complexity of planning algorithms since they do not need to find a full path in the geometrical space but a sequence of places to visit in a graph \cite{Luo_2018, Wiyatno_2022}.

\subsection{Simultaneous Localization and Mapping -  SLAM}

Although ego-motion methods have proven to be sufficient for pose estimation, their use for continuous localization does not solve the drift and error accumulation problems for long trajectories. Moreover, fine mapping is not possible without a continuous and robust pose estimation. While a robot navigates around an unexplored environment, the relative positions of the observed landmarks are all correlated due to the error of the robot's estimated location. This would make the problem intractable since for every new landmark the robot would need a new state with the robot position and the relationship of this new landmark to all the previous ones. It was not until 1995, with the new works on the convergence of the combined estimation of mapping and localization \cite{csorba1996new}, that authors realized that the optimization was feasible. A large number of landmarks is actually crucial to obtaining accurate solutions. 

Regarding localization, some approaches mainly focus on incrementally estimating the path of the agent, pose by pose, while performing local optimization. In the case of Simultaneously Localization and Mapping (SLAM), it aims at obtaining a globally consistent estimate of the agent trajectory while simultaneously mapping its surroundings \cite{Yousif_2015}, reducing drift errors. This is achieved by recognizing a location that has been previously visited and it is known as \textit{loop closure}. In other words, loop closure is the primary rectification step responsible for reducing the pose estimation drift errors when performing long trajectories, ensuring global consistency \cite{Tsintotas_2022}. Due to the sparse nature of the event streams, event-based approaches have the potential to achieve a significant reduction in computational complexity compared to current visual SLAM algorithms using conventional cameras. 

Next, the most relevant brain-inspired SLAM approaches are presented. In \cite{Weikersdorfer_vslam}, the authors presented vSLAM, an event-based 2D-SLAM method, as an extension of a previous event-based tracking formulation that adapts a particle filter condensation algorithm for self-localization \cite{Weikersdorfer_2012}. With every new event, the tracking algorithm uses a set of particles for a multivariate probabilistic estimate of the current system state. Each particle conveys the corresponding state estimation and its likelihood score. The map for tracking features is dynamically updated during localization and built as an occupancy map with each pixel representing the likelihood of events being triggered at its coordinates. Although the method reaches good accuracy, the solution is limited to 2D navigation and high-contrast scenarios. Moreover, there is no component for global consistency loop-closure. Lately, the same authors presented another extension for 3D-SLAM, but this time using a D-eDVS, a combined event-based-3D sensor (DVS + RGB-D) \cite{Weikersdorfer_2014}. Although this approach presented good results for 3D scenarios, the combination with the depth sensor limits the latency of the event camera. Again, a global optimizer to handle loop closure is still not integrated. 

The first truly event-based SLAM system ensuring loop closure was proposed by \cite{Milford2015TowardsVS}. It is considered a breakthrough in adapting current algorithms to work with event cameras. This naive approach consists of three key modules: 1) a loop closure algorithm built over the results of SeqSLAM \cite{Milford_2012}; 2) a limited visual odometry method constrained to the ground plane; 3) a semi-metric, topologically correct mapping tool based on the experience mapping algorithm \cite{Milford_2008}. For landmark recognition, SeqSLAM needs frame sequences. To adapt it for an event camera, events are temporally accumulated to create synthetic images. Finally, loop closure is integrated with the localization in the experience map as explained in \cite{Milford_2014}, showing reliable place recognition in diverse indoor and outdoor scenarios and navigating at different speeds. Authors in \cite{Nong_2022} proposed ASL-SLAM, an event-based visual SLAM approach that tracks features and event lines simultaneously. This approach showed improvement especially when operating under the presence of motion blur or low-frequency textures and under different lighting conditions. Its pipeline is similar to the PL-SLAM \cite{Pumarola_2017}, but uses synchronously extracted event lines combined with frame-based point feature correspondences. Following the same idea, in \cite{Chamorro_2022} authors built an approach that is based on the minimization of the event-line reprojection error. The method uses an error-state Kalman filter formulated entirely with Lie theory. Again, none of these approaches implement global loop closure. Finally, in \cite{wang2021asynchronous} authors proposed a method for an asynchronous Kalman Filter using an event sensor. The method is demonstrated in several applications including the deblurring of conventional frames when aggressive motion occurs. Recently, \cite{Fischer_2022} presented an event-based visual place recognition (VPR) algorithm that, given a new observation, retrieves the best matching reference from a database. This event-based VPR represents a valuable step towards loop closure on event-based SLAM algorithms. 

\subsection{Planning and actuation}
In robotics, planning methods split any high-level task into a sequence of low-level specific actions to accomplish it. For navigation, these actions are mainly the movements that the robot needs to perform to reach the final goal of the task. Most methods consider three areas in robotics planning: task planning \cite{Jiang_2018} that generates a sequence of actions; motion planning \cite{Chang_2020} that generates a sequence of movements; and trajectory/path planning \cite{Shao_2021} that produces a trajectory to reach the goal. For a detailed analysis of path planning methods, we encourage the reader to check \cite{Karaman_2011, Karur_2021}.

For the sake of completeness, this section briefly introduces the most popular path-planning approaches that are also the baselines for most bio-inspired works in the literature.     

\subsubsection{Path planning: local and global}
Path planning algorithms are used in autonomous robotic navigation to analyze and create trajectories from the robot's current location, or \textit{origin}, to the destination, which is referred to as \textit{goal}. Depending on the considered criteria, these path planning approaches optimize qualities such as traversed distance, time, battery level, or velocity, to achieve safe, collision-free, or efficient and least-cost paths. 

Planning considers local and global strategies. In fact a robust autonomous robot integrates both. \textit{Global planning} consists of finding a set of way-points that, when linked, describe a unique trajectory between the origin and the goal of the mission. Each way-point encodes a position-configuration change (maneuvers) or specific high-level action (such as taking a photo or measuring a magnitude). These planners normally do not take into account the robot's kinematic model. In contrast, \textit{Local planning} finds a trajectory between two way-points, avoiding obstacles. Global planning is done offline before any movement is executed and thus, it is risky since the current map may be imperfect or incomplete. Local mapping reduces risks locally, for example by updating the map with unexpected information or avoiding dynamic obstacles. Next, the two most well-known planners in the field are briefly described.     

\textbf{Rapidly-Exploring Random Trees (RRT)}. This algorithm \cite{LaValle_1998} is an incremental sampling-based planner that randomly samples and analyzes the map until a connection between origin and goal is found. This feature reduces computing time when planning paths, compared to other approaches that look for them in dense space. The algorithm works as follows: 1) The origin of the robot is set as the \textit{seed} of the \textit{tree}, which is a graph. 2) A random configuration (position) in the space is sampled and then, a collision detection algorithm validates it and checks for a free connection to the closest point of the current tree. 3) This process is iterated until a new node is close enough to the desired goal, outputting the path from the origin to this last node. Although it always guarantees at least one possible path in a limited time, sampling is random, and therefore, the path is sub-optimal. Finally, RRT can be used as a global or local planner. Some variations include RRT* that finds optimal paths \citep{Karaman_2011}. However, RRT* is computationally expensive, which makes it less suitable for onboard execution. FAST-RRT* \citep{Novo_2022} is a hybrid pseudo-optimal solution that combines RRT and RRT*, finding a more optimal path than RRT and faster than RRT*. The method uses a local-neighborhood tree reconnection of the path using a cost-based function.

\textbf{Probabilistic RoadMaps (PRM)}. In this case, the planner builds a graph of the map that will be used for future plans \cite{Kavraki_1996}. In contrast to RRT which has to sample and analyze the whole map for every requested plan, the PRM algorithm only needs to do it once for the entire exploration. When the map is sampled as a graph-like representation, better known as \textit{roadmap}, paths are found using graph-searching algorithms that considerably save computing resources. A roadmap is a connected graph in which nodes correspond to robot configurations (ie. positions in the space), and edges are simple trajectories between these nodes. The algorithm works as follows: 1) The configuration space is probabilistically sampled to create free nodes, providing a rather uniform covering of the space. 2) For each new node, a few neighbors in the graph are selected, and a local planner tries to connect them. 3) When this \textit{construction} phase is over, a second \textit{expansion} step is executed to refine the current graph and ensure its connectivity. 4) When a path is requested, two more nodes are added to the roadmap, if possible: the origin and the goal. Finally, the path is discovered by a graph-search algorithm. Popular graph-searching algorithms like Dijkstra 
, A* 
can be used to find these paths for the resulting roadmap. Some popular variants of this algorithm are PRM* \cite{Karaman_2011} and Lazy-PRM \cite{Bohlin_2000}.

\subsubsection{IMO Detection}
\label{subsubsec:IMO}
When the robot navigates the space, all points in the image are with respect to the perspective of the robot. Changes in the image can be due to the movement of the camera itself (ego-motion) or to the actual movement of objects in the scene (IMOs). The goal of IMO detection is to segment moving objects in the sequence while the robot is exploring. This can be difficult to accomplish because it is hard to estimate when the whole scene seems to be moving, due to the self-motion. Additionally, IMO detection usually involves the estimation of the pose and velocity of these dynamic objects. Finally, IMO segmentation approaches can be categorized \cite{Meunier_2022} into those focusing on the segmentation of a single object that is moving in the scene 
, and those that perform motion segmentation for multiple objects with a moving camera.

There are several ways to segment independently moving objects. Early approaches in the literature tried to subtract the overall camera motion from the estimated optical flow to segment the object. These techniques, in brain-inspired literature, are commonly referred to as \textit{motion compensation}. Then, more sophisticated approaches use ego-motion, projecting the 3D motion in the scene into the 2D camera plane, and iteratively subtracting it from the estimated motion to segment areas with residual motion until convergence. Finally, most recent approaches directly apply Deep Learning techniques using supervised and unsupervised methods. 

In \cite{Mitrokhin_2018}, authors proposed a computational graphic approach on the 3D point cloud of spatiotemporal events for ego-motion estimation. In their real-time pipeline, time windows are selected and then a motion compensation algorithm is applied to estimate the ego-motion of the camera, by fitting a parametric motion model. After ego-motion estimation, method continues with the segmentation of IMOs by clustering for the motion residuals. 0-MMS is an approach that also takes advantage of monocular motion compensation, and was presented in \cite{Parameshwara_2020}. The authors extracted event-based features and applied them for motion compensation while tracking them, yielding an approach that can accurately segment multiple independent moving objects in the scene, by following a strategy based on motion clusters. 

Authors in \cite{Alonso_2018} presented a CNN encoder-decoder for semantic segmentation using event cameras, with an encoder inspired by the \textit{Xception model} \cite{Chollet_2016} and a light decoder. The work compares different alternatives and proves the advantage of hybrid approaches that combine images and events.  In \cite{Mitrokhin_2019} the authors used two CNNs based on \cite{ye2020unsupervised}. The first neural network was devoted to estimating the depth of the scene by means of an event window. The second neural network was designed to estimate the position and orientation of each pixel using several event windows. Finally, by combining these outputs, optical flow and segmented IMOs are obtained. A key of this study is that they trained the shadow network with just 40k parameters and introduced a new dataset called EV-IMO for IMOs. Also, they mentioned that this architecture can generalize in a variety of scenes.

\begin{figure}
\centering
    \includegraphics[width=0.9\textwidth]{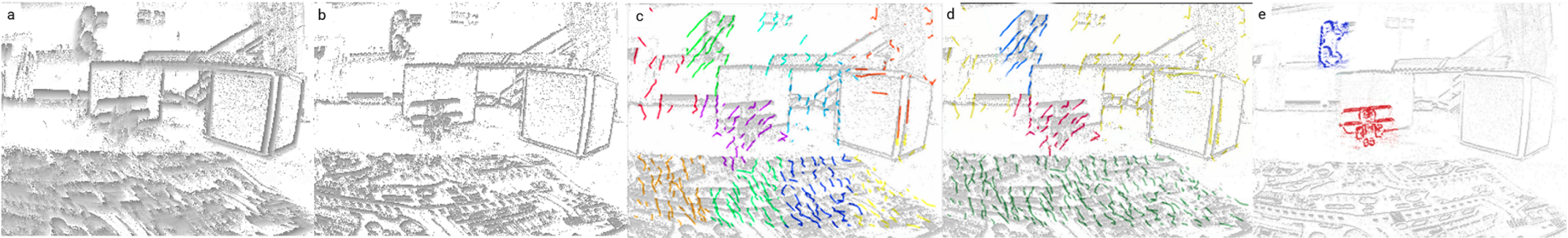}
    \vspace{-2.5mm}
    \caption{Overview of the proposed pipeline in \cite{Parameshwara_2020}. (a) Projection of the raw event cloud without motion
compensation, (b) Projection of event cloud after global motion compensation, (c) Sparse tracklets extracted on compensated event
cloud, (d) Merged feature clusters based on contrast and distance metrics, (e) Output of the pipeline is the cluster of events. The cluster membership is color-coded where the gray color indicates the background cluster (extracted from \cite{Parameshwara_2020}).}
    \label{fig:0-MMS}
    \vspace{-2.5mm}
\end{figure}

Finally, there is also an interesting method that applies SNNs to the problem of segmenting IMOs. SpikeMS \cite{Parameshwara_2021} is a fully end-to-end SNN for motion segmentation that employs an encoder-decoder architecture based on \cite{Mitrokhin_2020}, but feeding the events directly into the network without the need to use synchronized temporal windows. They obtained a good result but without remarkable improvement over those obtained with other methods.

\textbf{Datasets}. The most widely used datasets for IMO detection are EV-IMO \cite{Mitrokhin_2019} and EV-IMO2 \cite{Burner_2022}. However, previous attempts also achieved great popularity 
MOD \cite{Sanket_2019} and MOD++ \cite{Parameshwara_2020}. 

\textit{MOD} was originally intended for dynamic obstacle avoidance, attempting to avoid the pitfalls of other datasets that were built assuming static scenes \cite{Zhu_2018_dataset}. It contains 6 synthetic sequences that represent real-world indoor and outdoor scenarios. Every sequence has several light sources to recreate different illumination conditions with the camera moving around. The objects in the scene have been chosen using random textures with different shapes such as a ball, a drone, or a car. MOD++ extends MOD with additional sequences. \textit{EV-IMO2} extends EV-IMO with more complex scenarios, including RGB images, and increasing the number of sensors providing data with 2 Prophesee Gen3, a Samsung Gen3, and a conventional Flea3 camera, and a MoCap system to obtain the ground truth. The dataset contains 173 indoor sequences grouped into three categories: independently moving household objects, static scenes, and basic motions in shallow scenes. Some of them were recorded in low light conditions. There are over 20 different objects such as small drones, toy blocks, or remote control cars. These datasets are summarized in Table \ref{tab:eventdatasets}.

\subsubsection{Obstacle avoidance}
When navigating in any environment, the robot needs to reach the destination by dodging any unexpected static or dynamic obstacle. This task is particularly crucial when the robot navigates a dynamic environment and needs to act as fast as possible to avoid collision with a static obstacle or even another mobile robot. Challenges for obstacle avoidance include cluttered environments or fast objects that require very precise and high-speed maneuvers. Thus, reducing the perception latency for high-speed obstacle avoidance has become a challenge in autonomous robotics \cite{Barry_2018, Sai_2021}. Precisely, event-based sensors have become the best candidates for obstacle avoidance given their very low latency.  

\cite{Falanga_2019} presented a study analyzing, from a theoretical perspective, the perception latency vs. actuation agility trade-off at maximum speed, to navigate autonomously in a static scenario. In their experiments, an event camera is used for dynamic obstacle avoidance on a quadrotor platform. Event cameras allow the robot to achieve a speed increase between 7\% and 12\% when compared with conventional cameras. EVDodgeNet \cite{Sanket_2019} is a deep learning network that performs multiple dynamic obstacle avoidance using event cameras. The method uses the inertial inputs from the platform and the event stream, which is stabilized using motion-compensated by using a neural network. The stabilized input is then processed by two additional networks that perform ego-motion and IMO detection estimation. All the results are used as a control policy for dodging multiple dynamic obstacles. Another interesting application for obstacle avoidance in low lighting conditions using event cameras was proposed in \cite{Yasin_2020}. The pipeline has four stages: 1) background event activity noise cancellation; 2) object detection module using an adaptive slicing algorithm and a Hough transform \cite{XU1993131} and in parallel, a module that performs 2D-corner detection \cite{Vasco_2016}; 3) depth estimation of the 2D features; 4) estimation of the relative velocity of the object to avoid collision. 

\subsubsection{Tracking}
Tracking an object starts with its detection is why some works referenced in the IMO section also include tracking. Here, we will discuss works focused solely on tracking.

Tracking aims to detect and follow an object during the action and is needed in planning and obstacle avoidance. But it can also be used for object tracking alone. Most tracking approaches are mainly linked to the detection of IMOs, their classification, and labeling for a certain time, depending on the required task. Although these approaches are computationally intensive, there are implementations with neural networks that run in real time thanks to event cameras. For example, \cite{Zhang_2022_CVPR} proposes a spiking transformer network. The architecture is based on SNN and uses two components. The first is the SNNformer Feature Extractor (SFE), which takes advantage of the Swin transformers \cite{Liu_2021_transformers} to extract global spatial features from events. The second component is temporal-spatial feature fusion (TSFF). This method runs in real-time outperforming the state-of-the-art but, it only tracks a single object.

\section{Spatial-cell Models for Cognitive Autonomous Navigation}
\label{subsec:brain_spatialcells}
Apart from the models for isolated components, holistic approaches for autonomous navigation that mimic different brain areas and structures involved in navigation have been proposed. The goal is to take advantage of neuroscientific developments to formulate new effective lightweight solutions deployable in small and low-power computing platforms.

As mentioned in Section \ref{subsec:simulation}, although other brain areas do play a role in navigation, the entorhinal cortex and the hippocampus are the main ones in charge of building a cognitive map of the scene and supporting localization within it. Simplifying the roles of the specialized cells that are localized in these brain regions: PCs could be seen as spatial units for mapping external features onto internal and local metric coordinates in a distributed cognitive map; HDCs could be conceived as some kind of internal compass; and GCs represent the position within this map. Furthermore, space representation is a process done in an idiothetic manner. Over time, these brain areas integrate self-motion cues from different sensory inputs to estimate pose and location relative to a reference initial point, the so-called path integration. Next, we introduce computational models for the most relevant autonomous navigation works that mimic the working principles of these brain regions.

\subsection{Computational models for spatial cells}
\label{subsec:models_cells}
Different computational models of the PCs firing rate have been proposed based on the results of biological experiments on rats. \cite{Hartley_2000} modeled the inputs to the PCs as a population of Boundary Vector Cells (BVC), each of which fires maximally when a boundary is at a particular distance and allocentric direction to the agent \cite{Barry_2006, Lever_2009}. Thus, the PCs firing rate is modeled as proportional to the thresholded sum of the BVC inputs that it receives by using a threshold-linear transfer function. The model is able to describe PCs adaptation during learning, such as the consolidation of place fields and removal of duplicate fields. However, it does not perform well when map cues are removed, missing the remapping feature of PCs. More recently, \cite{Mazzara_2022} introduced a detailed morphological and biophysical model implementation for the creation of PCs. 

Regarding GCs and HDCs, for example, \cite{Kropff_2008} presented a model of grid field formation, where the grid emerges from the contrast between the space continuity and the neuronal fatigue firing. The grid is built as the result of Hebbian learning in the feed-forward connections of their model. Different oscillatory-interference models have been also proposed to encode the speed and direction of motion information into a grid pattern in the frequency of the membrane-potential oscillators \cite{geisler2007hippocampal, traub2022simulation}, or the connections between the oscillations and the stability of the grid representations \cite{koenig2011spatial, yan2021model}. Studies performed afterward also showed that GCs are modulated by local directional signals \cite{gerlei2020grid}.  

\subsection{Full navigation models}
\label{subsec:full_nav}
Some works in the literature are based on the computational models proposed for the specialized cells presented in Section \ref{subsec:models_cells}, building structures that replicate the functioning of brain regions for navigation. For example, in \cite{Li_2017} the authors proposed a method for localization based on PCs. The whole processing pipeline consists of a first stage that detects and assigns landmarks building an internal map of PCs that represents the scenario, and a second stage that represents the relationships between the distances and the PCs' firing, using a Radial Basis Function (RBF) neural network \cite{Alfa_2016}. Another model that combines PCs and GCs, applied to goal navigation learning was presented by \cite{Yan_2016}. The interest of this work relies on how a simulated agent is capable of, after learning environmental cues during several executions, finding a path to the aim by optimizing it at every iteration, merging the PCs and GCs learning behavior. The learning rule of the model consists of a winner-takes-all mechanism, where the cell with the maximal firing rate at a particular position wins the learning chance \cite{kulvicius_2008}. Similarly, \citep{Chen2019} presented another goal-oriented brain inspired navigation model. In this work, the robot is able to converge to an optimal path by the guidance of population activities and reward signals. A grid-like representation for flexible navigation in challenging environments combined with theoretical models of GCs and Vector-based Navigation algorithms was proposed in \citep{Banino2018}. They also showed how their algorithm was able to exploit optimal routes when combined with a path-based barrier avoidance strategy.

Other approaches propose models that better integrate the different cells such as \cite{Han_2020}, with a model for PC-GC generation that could be used as the foundations for the construction of cognitive maps. In this work, PCs and GCs generated from environment visual information are called \textit{visual-PCs} and \textit{visual-GCs} respectively. Then, the fusion of the \textit{visual-GCs} with self-motion cues is called \textit{united-GCs}, while PCs generated via supervised learning are called \textit{united-PCs}. The first stage of the processing pipeline is the simulation of \textit{visual-PCs} using a Gaussian function on the absolute location information; new cells are created during the exploration process every time the firing of the last activated cell goes below a fixed threshold. The firing rate depends on the distance from the explored location to the boundary. In the second stage, these cells generate the \textit{visual-GCs} through a feed-forward network \cite{Castro_2014} to build their hexagonal firing pattern distribution. Next, in the third stage, \textit{visual-GCs} are combined with self-motion cues using a genetic algorithm \cite{Katoch_2021}, thus generating \textit{united-GCs}. In previous steps, since cells only use the boundary information as input, GCs cannot expand to the whole space. However, when including the self-motion information, the firing field can now be extended with additional information, not limited by the boundaries. Finally, in the fourth stage, \textit{united-PCs} are built by combining \textit{united-GCs} and \textit{visual-PCs} to improve the position accuracy of PCs away from the scene boundaries. This is done via a supervised fuzzy ART network \cite{Carpenter_1991}, which is a competitive classifying and clustering network that implements plasticity and enables \textit{continual learning}, so it has the ability to learn new knowledge without suffering from \textit{catastrophic forgetting}.

Another set of models combined PCs and GCs with HDCs, improving tasks such as collision avoidance while navigating. Such a naive bio-inspired model was presented by \cite{Gay_2021, Gay2023}. In this work, PCs are modeled so that the environment is encoded in \textit{contexts} that contain currently observed visual cues given in polar coordinates. The PC associated with the current position fires by comparing the similarity of the current context to previous ones. For each context, a set of HDCs is created for the virtual rotations of the current context at a fixed angle. GCs are also modeled in a similar way to PCs and HDCs, by the similarity of displaced context, but in a specific activated PC context. GCs and HDCs contexts are used in combination to estimate the orientation and position of the robot. The model was validated on basic functions like mapping, guidance, homing, and shortcut finding, although is carried out in very simple scenarios. 

Finally, some approaches also proposed models of the Lateral Entorhinal Cortex (LEC) and the Medial Entorhinal Cortex (MEC), regions that produce inputs for the hippocampus. In \cite{Zhao_2021}, the authors proposed a hierarchical network mimicking the LEC and MEC cortical areas for robot position prediction. In the proposed neural architecture the system learns the transformation of egocentric sensorimotor inputs into allocentric spatial representation for navigation, using a deep learning approach. The feed-forward connections, which convey spatiotemporal information, are learned by stochastic gradient descent. More recently, the same authors presented SeMINet \cite{Zhao_2022}, with a model that builds the map integrating self-motion and visual features. 

\subsubsection{Continuous Attractor Neural Networks (CANNs)}

A special category of approaches for navigation is based on CANNs \cite{Samsonovich_1997, Wu_2016}. These networks are dynamic models of cells imitating the behavior of the navigation neurons already mentioned in Section \ref{subsec:models_cells} \cite{Knierim_2012}. They are built on cells connected to nearby cells by excitatory (positive) or reinforcing connections that allow energy in the central cell to propagate. Also, there are inhibitory (negative) connections from each cell to every other single cell in the network. With this inhibition, if a cell is activated, for example, a cell representing a specific orientation, it inhibits the activity in all the other cells representing other directions in the network by a certain amount. CANNs are considered to be a canon for the GCs modeling. The excitation between cells is usually represented via a Gaussian distribution. Following the formulation by \cite{Milford2003HippocampalMF}, in a 2D-CANN of dimensions $(n_x,n_y)$, the excitation weights are computed as shown in Eq. \ref{eq:excitation_cann}. In this expression, $\delta_{x}$ and $\delta_{y}$ are constants of the variance for the 2D spatial distribution, and the distance between units cells is encoded by $u=(x-i)(mod~n_{x})$ and $v=(y-j)(mod~n_{y})$.
\begin{equation}
\varepsilon _{u,v}^{gc}=\frac{1}{\delta_{x}\sqrt{2\pi}}e^{\frac{-u^2}{2\delta_{x}^2}}\cdot \frac{1}{\delta_{y}\sqrt{2\pi}}e^{\frac{-v^2}{2\delta_{y}^2}}
\label{eq:excitation_cann}
\end{equation}
Thus, for each cell $(x,y)$, the total excitation from every cell $(i,j)$ is expressed as shown in Eq. \ref{eq:excitation_cann_incr}.
\begin{equation}
\Delta P_{\textbf{x}}^{gc}=\sum_{i}^{n_x}\sum_{j}^{n_y}P_{i,j}^{gc}\cdot \varepsilon _{u,v}^{gc}
\label{eq:excitation_cann_incr}
\end{equation}
And next, local and global inhibition is obtained as in Eq. \ref{eq:inhibition_cann_incr}. 
\begin{equation}
\Delta P_{\textbf{x}}^{gc}=\sum_{i}^{n_x}\sum_{j}^{n_y}P_{i,j}^{gc}\cdot \psi  _{u,v}^{gc}-\varphi
\label{eq:inhibition_cann_incr}
\end{equation}
Where $\psi _{u,v}^{gc}$ is an inhibitory weight matrix to update local inhibition activity, and $\varphi$ is a constant to perform global inhibition equally to all the cells' activity, which must not be negative. And finally, the total activity in the CANN is normalized as shown in Eq. \ref{eq:norm_cann_incr}.
\begin{equation}
P_{\textbf{x}}^{gc'}=\frac{P_{\textbf{x}}^{gc}}{\sum_{i}^{n_x}\sum_{j}^{n_y}P_{i,j}^{gc}}
\label{eq:norm_cann_incr}
\end{equation}
Many models in the literature use CANNs and some of the most popular were conceived by mimicking the neural behaviour underlying the navigation process in the rodent brain. The open-source navigation RatSLAM model \cite{Ball_2013} is inspired by the rodent's hippocampus and estimates position and orientation using visual cues from a single low-resolution camera. GCs are simulated by what they call a \textit{Pose Cell Network}, which is indeed a 3D-CANN that encodes the pose of the agent given by $(x,y,yaw)$. These cells refine the pose estimation using previous computations but in a smooth and continuous transitional manner. The input to this network comes from two separate sources: 1) translational and rotational velocity from a simple visual odometry algorithm; 2) responses from \textit{Local View Cells}, related to the PCs. These cells learn, for every new scenario, the activity of the \textit{Pose Cell Network} at that moment, and inject this energy again to the network when revisiting a known place, a feature that is crucial for robot relocalization. Finally, loop-closure is performed using an experience map \cite{Milford_2014}. 
As proof of the potential of RatSLAM, it was able to successfully map an entire neighborhood, or to navigate in an office environment for over two weeks \cite{Milford_2008, Milford_office}.

NeuroSLAM \cite{Fangwen_2019} is a 4-DoF SLAM method built on GC and HDC models, integrated with a vision system that provides external visual cues and self-motion cues. This new model is inspired by 2D RatSLAM but extends it to 3D mapping and localization. In \cite{Chen_2021}, the authors also introduced a model of PCs and HDCs based on CANNs to encode visual and inertial inputs, and then continuously compute the current orientation and position. 

Biological path integration was originally modeled using vector navigation \cite{Edvardsen_2015, Bush2015UsingGC}. In \cite{Edvardsen_2017}, the author proposed a GC-based CANN navigation system, since GCs are thought to be part of the path integration processes and to be responsible for building the coordinate system of the environment. The authors used CANNs that were able to perform path integration by using the inputs from GCs. However, vector navigation is also interesting, pointing to the direction from the current location to a specific goal or a previously visited location stored in memory. This is usually referred to as the goal-direction signal, and it is crucial for a robot navigating the real world. An improvement in long-range navigation was presented later in \cite{Edvardsen_2019}, with models that allow robots to navigate up to 100 meters in the experiments. More recently, the same author extended this work by integrating PCs and BVCs for navigation and local obstacle avoidance \cite{Edvardsen_2020}. In this work, the authors highlight that while PCs are linked to a topological representation of the environment, GCs are thought to support metric navigation. Therefore, GCs alone are not enough for navigation in cluttered environments, where obstacle avoidance is required. Even when integrating BVCs, the robot may get stuck and PCs will be required to undo some of the previous steps and attempt a new trajectory, using what authors called an experience-dependent topological graph.      

Finally, an applied work \cite{Chao_2021}, shows a successful system that uses GCs and HDCs to model a localization method for Unmanned Aerial Vehicles (UAVs). HDCs are built into a 1D-CANN to model the yaw rotational velocity of the UAV and this result is fed to a 3D-CANN built with GCs. This second 3D-CANN estimates the UAV translational velocity, which is provided to a last stage that decodes position from the firings of GCs by using a period-adic method. The method is computationally simple, and experiments show robust estimations under noisy conditions. 

\section{Discussion}
\label{sec:discussion}
Event cameras are rapidly gaining popularity as a promising neuromorphic technology. They offer several advantages over frame cameras, such as low power consumption, low latency, and high dynamic range. With advances in control policies, agile robots can make aggressive maneuvers, but their perception and processing pipelines are limited by latency. However, event cameras have opened up new possibilities to push these limits further. They can efficiently compress visual information by suppressing redundant data, thereby reducing processing pipeline latency. This intrinsic ability of event cameras makes it possible to close perception-action loops in real-time specifically when using event-wise processing that exploits the asynchronous sparse nature of the event streams. SNNs have the potential to become the best-suited networks for event-driven processing, as these neural architectures have shown to be capable not only of processing asynchronous data streams but also of processing more efficiently than conventional networks. Moreover, SNNs are able to adapt their connections to the input over time, without any explicit reprogramming, a cognitive process called plasticity. 

Regarding the computational models of areas of the brain involved in navigation, they are also gaining popularity since biology has been shown to efficiently perform navigation. The key aspect underlying this energy-efficient processing comes from the environment representation that is built and how this cognitive representation is interpreted by specialized navigation cells to determine the location within it. This type of representation is believed to be responsible for the low-power consumption of the brain when perceiving, reasoning, and acting during the navigation process; thus, it is valuable research for mobile robotics. To this end, this paper discusses future directions of neuromorphic engineering:

\textbf{Exploiting the temporal information of event cameras}: It can be noticed that most current solutions try to readapt the traditional computer vision algorithms by using synthetic virtual frames as input, aggregating events. The survey shows that there are better approaches to handling event data because temporal information is partially or completely ignored. Some other approaches have tried to redesign this paradigm representation, proposing alternatives such as volumes of events, time surfaces, etc. However, there has yet to be an established event representation or processing technique. Indeed, the works that present better results are those that use a combination of different sensors. This means that there is still much work to be done with these novel cameras. 

\textbf{ANNs vs. SNNs}: Regarding the state-of-the-art methodology for event cameras, on the one hand, for most applications, deep learning methods lead the way. For example, neural networks offer more potential to infer dense optical flow than traditional methods. This is because their architecture is based on multi-scale models, which can abstract global and local scene information using a 3D representation of events or even directly using the stream of events. On the other hand, model-based methods just started to emerge for ego-motion estimation. However, these architectures are purely von Neumann, which means they take advantage of grid-like representations given synchronously at a fixed time interval; new mechanisms that exploit the asynchronous nature of the streams of events are required.
Our review shows that SNNs still lack accuracy compared to ANNs. However, this gap is being reduced and in fact, some task-specific SNNs already perform better. The reader should keep in mind that SNNs provide efficient information encoding, leading to reduced computational complexity and energy consumption. Finally, SNNs are the way of asynchronously processing spikes or events when implemented in neuromorphic hardware. However, there is still a long way to go before scalable processing architectures and hardware can be used to solve real-world problems.

\textbf{Overcoming the limitations of CANNs}: The current state-of-the-art computational models that replicate brain areas focus on studying specialized navigation cells modeled by CANNs. CANN-based models have been shown to successfully perform navigation in unstructured environments in experiments in real scenarios. The problem with the CANNs is that they need to be more scalable and are used to solve simple problems such as direction estimation. Moreover, regarding SLAM, a bio-inspired method for loop closure is still needed to be successful.

\textbf{Compromise between energy and accuracy}: In terms of energy efficiency, neuromorphic computing has been demonstrated to outperform deep learning approaches. However, they still lack accuracy performance and energy efficiency is only attainable when implemented in neuromorphic computing platforms. This combination could raise neuromorphic computing to a new stage as the successor of current deep learning approaches. Combining neuromorphic sensory, computing platforms, and neural algorithms would mean a new step towards effective biologically-inspired robotics with lower latency, higher energy efficiency, and lower inference times.  

\textbf{Brain-inspired 3D navigation methods}: Many of the navigation approaches inspired by the brain spacial cells architecture tackle only the 2D SLAM, and most of them are very naive as the functional relationship between brain cells is still not entirely clear. We hope that recent discoveries of 3D spatial reasoning in the brain will lead to new and promising 3D brain-inspired SLAM, as presented by NeuroSLAM \cite{Fangwen_2019}.

\section{Conclusions}
\label{sec:conclusions}
This survey has covered research on bio-inspired methods for robotics applications, focusing on vision-based navigation tasks. First, we have provided an overview of the global pipeline in autonomous robotic systems, from perception to actuation. Although this survey has presented several brain-inspired perception approaches and sensors, it remains a great challenge to develop efficient end-to-end systems that also target control and actuation following bio-inspired principles, with promising advances in fields such as active vision that pairs perception and control. 

Bio-inspired techniques are gaining popularity in robotics due to long-standing evidence that animals operate more effectively than state-of-the-art navigation algorithms. We conclude that, despite recent bio-inspired breakthroughs, the paradigm of the perfect autonomous system is still a long way off from reality. But it is getting close, as neuromorphic sensors or brain-inspired computing platforms have shown us. Moreover, some navigation algorithms based on the brain cognitive representation and scene understanding, are coming out. Although naive, these solutions are the first steps toward models inspired by the human brain. Finally, in response to the initial question we presented in this paper \textit{``Can neuromorphic engineering be the successor of current and classical navigation solutions in robotics?''}, we anticipate that, eventually, the advantages of brain-inspired navigation algorithms will combine with the ones from the event-based sensors approaches, providing an innovative approach for effective bio-inspired autonomous robotics.

\paragraph{Acknowledgements} This work was supported by the Spanish National Grant PID2022-141466OB-I00 funded by MICIU/AEI/10.13039/501100011033 and by ERDF/EU. 
	
\bibliography{references}
	
\end{document}